\documentclass{article}

    \usepackage[preprint]{neurips_2024}

\usepackage{natbib}
\setcitestyle{numbers,square}
\usepackage[utf8]{inputenc} % allow utf-8 input
\usepackage[T1]{fontenc}    % use 8-bit T1 fonts
\usepackage{hyperref}       % hyperlinks
\usepackage{url}            % simple URL typesetting
\usepackage{booktabs}       % professional-quality tables
\usepackage{amsfonts}       % blackboard math symbols
\usepackage{nicefrac}       % compact symbols for 1/2, etc.
\usepackage{microtype}      % microtypography
\usepackage{xcolor}         % colors

\usepackage{amsmath}
\usepackage{mathtools}
\usepackage{amsthm}

\usepackage{CJK}
\usepackage{latexsym}

\usepackage{amssymb}
\usepackage{graphicx}
\usepackage{pdfpages}
\usepackage{enumitem}
\usepackage{amsmath}
\usepackage{url}
\usepackage{subfigure}
\usepackage{caption}
\usepackage{algorithm}
\usepackage{setspace}
\usepackage{algpseudocode}
\renewcommand{\algorithmicrequire}{\textbf{Input:}}
\renewcommand{\algorithmicensure}{\textbf{Output:}}
\theoremstyle{definition}
\newtheorem{definition}{Definition}[section]
\AtBeginDocument{%
  \providecommand\BibTeX{{%
    \normalfont B\kern-0.5em{\scshape i\kern-0.25em b}\kern-0.8em\TeX}}}

\algtext*{EndIf}

\title{Sora Detector: A Unified Hallucination Detection for Large Text-to-Video Models}

\author{%
  Zhixuan Chu$^{1}$, Lei Zhang$^{1}$, Yichen Sun$^{2}$, Siqiao Xue$^{1*}$, Zhibo Wang$^{2}$, Zhan Qin$^{2*}$, Kui Ren$^{2}$ \\
  $^1$ \text{Ant Group}  $^2$ \text{Zhejiang University} \\
  \texttt{\{chuzhixuan.czx,xiaobei.zl,siqiao.xsq\}@antgroup.com}\\
  \texttt{\{yichensun,zhibowang,qinzhan,kuiren\}@zju.edu.cn} \\
}

\begin{document}

\maketitle
\def\thefootnote{*}\footnotetext{Corresponding author.}

\begin{abstract}
The rapid advancement in text-to-video (T2V) generative models has enabled the synthesis of high-fidelity video content guided by textual descriptions. Despite this significant progress, these models are often susceptible to hallucination—generating contents that contradict the input text—which poses a challenge to their reliability and practical deployment. To address this critical issue, we introduce the \textbf{SoraDetector}, a novel unified framework designed to detect hallucinations across diverse large T2V models, including the cutting-edge Sora model. Our framework is built upon a comprehensive analysis of hallucination phenomena, categorizing them based on their manifestation in the video content. Leveraging the state-of-the-art keyframe extraction techniques and multimodal large language models, SoraDetector first evaluates the consistency between extracted video content summary and textual prompts, then constructs static and dynamic knowledge graphs (KGs) from frames to detect hallucination both in single frames and across frames. Sora Detector provides a robust and quantifiable measure of consistency, static and dynamic hallucination. In addition, we have developed the \textbf{Sora Detector Agent} to automate the hallucination detection process and generate a complete video quality report for each input video. Lastly, we present a novel meta-evaluation benchmark, \textbf{T2VHaluBench}, meticulously crafted to facilitate the evaluation of advancements in T2V hallucination detection. Through extensive experiments on videos generated by Sora and other large T2V models, we demonstrate the efficacy of our approach in accurately detecting hallucinations. The code and dataset can be accessed via \href{https://github.com/TruthAI-Lab/SoraDetector}{https://github.com/TruthAI-Lab/SoraDetector}.
  
\end{abstract}

\section{Introduction}

The rapid advancement of large language models (LLMs) has revolutionized various domains, including intelligent dialog systems~\citep{nakano2021webgpt,xue2023weaverbird}, time series forecasting \cite{jin2023time,xue2023prompt}, recommendation systems~\citep{chu2023leveraging}, database applications~\citep{langchain,xue2023dbgpt,xue2024demonstration} and multimodal generations~\citep{radford2021learning}. Among the breakthroughs, the emergence of large text-to-video (T2V) language models~\citep{wu2023tune,singh2023survey} has garnered significant attention due to their ability to generate realistic and imaginative videos from textual descriptions. These models have the potential to transform creative industries, entertainment, education, and scientific visualization by enabling the creation of engaging and immersive video content. One notable development in this field is the introduction of Sora~\citep{sora}, an upcoming generative artificial intelligence model developed by OpenAI. Sora specializes in T2V generation, accepting textual descriptions, known as prompts, from users and generating short video clips corresponding to those descriptions. The model's versatility allows it to generate videos based on artistic styles, fantastical imagery, or real-world scenarios, showcasing its ability to understand and interpret a wide range of concepts.

However, the development and deployment of large T2V language models like Sora also raise important considerations, particularly in terms of hallucination detection. Hallucinations ~\citep{chen2024factchd} refer to the generation of content that is inconsistent, anomalous, or diverges from the intended description. These inconsistencies can manifest in various forms, such as objects appearing or behaving implausibly, inconsistencies in visual attributes, or deviations from the specified context. Detecting hallucinations in videos~\citep{huang2023opera,chen2024unified} generated by large T2V language models presents distinct challenges compared to hallucination detection in text or image generation. Videos are inherently temporal and dynamic, requiring the consideration of consistency and continuity across multiple frames. Traditional static hallucination detection methods, which focus on anomalies within individual frames, fall short of capturing temporal inconsistencies and anomalies that emerge over time. Detecting such anomalies necessitates a holistic approach that takes into account the temporal evolution of the video content.

To address these challenges, we introduce a comprehensive methodology designed specifically for identifying hallucinations in videos generated by large T2V language models like Sora. Our approach combines multiple techniques, including keyframe extraction~\citep{guan2012keypoint}, object detection~\citep{li2023evaluating}, knowledge graph construction~\citep{wang2023enhancing}, and multimodal large language models~\citep{gpt4}, to detect both static and dynamic hallucinations effectively. In addition, to advance research in this intricate field and facilitate the development and evaluation of video hallucination detection methods, we introduce a comprehensive (\textbf{T2V}) \textbf{Ha}l\textbf{lu}cination Detection \textbf{Bench}mark, \textbf{T2VHaluBench}, specifically designed for assessing the performance of hallucination detection algorithms in large T2V language models. This dataset serves as a valuable resource for researchers and practitioners, enabling them to assess the effectiveness of their hallucination detection approaches and foster further advancements in this field.

Our experimental results demonstrate the effectiveness of our methodology in detecting hallucinations across a wide range of video generation tasks. The proposed approach consistently outperforms baseline methods, achieving significant improvements in precision, recall, and F1 score. The ablation studies conducted on individual components of our methodology highlight the importance of each detection stage and the synergistic effects of combining them. In summary, our main contributions are:
\begin{itemize}[leftmargin=*]
   \item We establish a comprehensive framework, \textbf{SoraDetector}, for detecting hallucinations in T2V generation, covering an extensive variety of hallucination categories. This approach enhances the overall comprehension of the issue across large T2V models, promoting a unified perspective on identifying and addressing these inconsistencies.

   \item We introduce the \textbf{Sora Detector Agent}, an automated hallucination detection system built upon large language models (LLMs). The Sora Detector Agent streamlines the hallucination detection process by integrating the Sora Detector framework with LLMs, enabling the generation of comprehensive video quality reports for each input video. This agent enhances the accessibility and usability of our hallucination detection methodology, making it more adaptable for real-world applications and user-friendly for researchers and practitioners alike.

   \item We unveil \textbf{T2VHaluBench}, a benchmark that encompasses various hallucination categories in T2V generations. This benchmark is equipped with fine-grained analytical features, gauging the progress of hallucination detectors.
\end{itemize}

\section{Background}
In this section, we provide an overview of the foundational background related to large text-to-image (T2I) and T2V models, upon which our work is predicated. In addition, we review the technical context of hallucination in the works of T2I and T2V.

\subsection{Text-to-image Generations}
Most of the common approaches for T2I generation are based on diffusion models~\citep{zhou2024survey}. Of these models, DALL-E2~\citep{ramesh2021zero}
and Imagen~\citep{saharia2022photorealistic} achieve photorealistic T2I generation using diffusion-based models. A promising line of works design text-to-image diffusion models that generate high-resolution images end-to-end, without a spatial super-resolution cascaded system or fixed pre-trained latent space~\citep{gu2023matryoshka}. This progress highlights the rapid advancement in research and its application in generating visually engaging outputs, with each
step marking significant leaps in the technical sophistication of image generation.

\subsection{Text-to-video Generations} 
The principal concept of T2V generation is to produce dynamic and contextually rich videos directly from textual descriptions. Recently, the diffusion models~\cite {wu2023tune,zhou2022magicvideo} become dominant approaches to producing dynamic and contextually accurate video content from textual
descriptions. Coupling with learning using large-scale datasets, they generate videos that are not only visually stunning but also maintain narrative coherence throughout. Each advancement in T2V generation models significantly deepens our understanding of the intricate relationship between textual narratives and visual storytelling.

\subsection{Hallucinations in Text-to-image Generations}
In general, hallucination refers to a
generative model imagines factually incorrect details in its response for a given input. In T2I tasks, hallucinations occur when the model imagines incorrect details about an image in visual question answering. Existing works on addressing hallucinations in T2I tasks include building open benchmarks~\citep{huang2024visual,li2023evaluating}, mitigating hallucinations by improving the fine-tuning data quality~\citep{wang2023mitigating} or model structure~\citep{tong2024eyes}.

\subsection{Hallucinations in Text-to-video Generations}

Recent research has begun to address the issue of object hallucination in T2V generations, including hallucination evaluation and
detection~\citep{wang2023evaluation,liu2024mitigating}, as well as the construction of higher-quality datasets for fine-tuning~\citep{gunjal2023detecting,li2023m}. Nonetheless, no end-to-end method has been proposed to address the hallucination for large T2V models. We have already demonstrated the effectiveness of our method in reducing hallucination and its compatibility with various T2V models.

\section{Related Work}
\subsection{KeyFrame Extraction}

In video key frame extraction, various methodologies have been proposed to tackle this task from different perspectives, each with its merits and shortcomings. 

Segmentation-based methods, for instance, hinge on detecting significant alterations in frame-to-frame similarity, often quantified by thresholds in image characteristics such as color histograms, edge detection, or motion vectors. Reference ~\citep{panagiotakis2009equivalent} discusses key frames that maintain equal Iso-Content Distance, Error, and Distortion; reference ~\citep{guan2012keypoint} introduces a selection process that taps into a global selection of SIFT feature-based key points across all frames. These methods' limitations are the potential extraction of duplicate key frames if certain content reappears within the video.

 In addition, dictionary-based methods treat key frame extraction as a sparse dictionary selection issue where techniques such as sparse dictionary selection using the $L_{2,1}$ norm and the exact $L_{0}$ norm constraint play a pivotal role, as discussed in ~\citep{cong2011towards} and ~\citep{mei2015video} respectively. Moreover, to augment these approaches, ~\citep{mei20142} suggests an $L_{2,0}$ norm-constrained model and ~\citep{wang2017representative} incorporates structured regularizers, striving for a balance between representativeness and diversity; however, these methods are complex. 

Clustering-based techniques are straightforward and yield easily interpretable results. They work by grouping similar frames and designating the frame nearest to each group's centroid as the key frame. Reference ~\citep{zhuang1998adaptive} employs unsupervised clustering to discern key frames based on the visual variation. More sophisticated variants such as Dynamic Delaunay clustering in ~\citep{kuanar2013video}, k-means clustering with color features in ~\citep{de2011vsumm}, spectral clustering on spatiotemporal features in ~\citep{vazquez2013spatio}, graph clustering as utilized in ~\citep{panda2014scalable}, and the use of local descriptors alongside graph modularity for extraction in ~\citep{gharbi2017key}.

\subsection{Knowledge Graph and LLM}

\textbf{Knowledge Graph.} Knowledge Graph (KGs) are structured multirelational knowledge bases that typically contain a set of facts. Each fact in a KG is stored in the form of triplet $(s, r, o)$, where $s$ and $o$ represent the subject and object entities, respectively, and $r$ denotes the relation connecting the subject and object entity. KGs are crucial for various applications as they offer accurate explicit knowledge~\citep{Ji_2022,sheu2021knowledge}. Besides, they are renowned for their symbolic reasoning ability~\citep{zhang2021neural}, which generates causal and interpretable results \citep{chu2024task,chu2021graph}. KGs can also actively evolve with new knowledge continuously added in. Additionally, experts can construct domain-specific KGs to provide precise and dependable domain-specific knowledge.

\noindent \textbf{LLM.} LLMs, pre-trained on the large-scale corpus, such as ChatGPT~\citep{gpt3} and GPT-4~\citep{gpt4} have showcased their remarkable capabilities in engaging in human-like communication and understanding complex queries, bringing a trend of incorporating LLMs in various fields~\citep{anil2023palm,gunasekar2023textbooks,chu2023data}. These models have been further enhanced by external tools, enabling them to search for relevant online information~\citep{nakano2021webgpt}, build recommendation systems~\citep{chu2024llm}, utilize tools~\citep{schick2023toolformer}, and create more sophisticated applications~\citep{langchain}. 

Despite their success in many applications, LLMs have been criticized for their lack of factual knowledge. Specifically, LLMs memorize facts and knowledge contained in the training corpus. However, further studies reveal that LLMs are not able to recall facts and often experience hallucinations by generating statements that are factually incorrect, i.e., hallucination~\citep{Ji_2023}.

\noindent \textbf{KG-enhanced LLMs.} Integrating KGs can enhance the performance and interpretability of LLMs in various downstream tasks~\citep{Pan_2024}. KGs store enormous knowledge in an explicit and structured way, which can be used to enhance the knowledge awareness of LLMs \citep{wan2024bridging}. In our work, based on the knowledge retrieved from KGs, we further validate whether the factual knowledge is hallucinated.

\begin{figure*}
  \centering
    \includegraphics[width=1\linewidth]{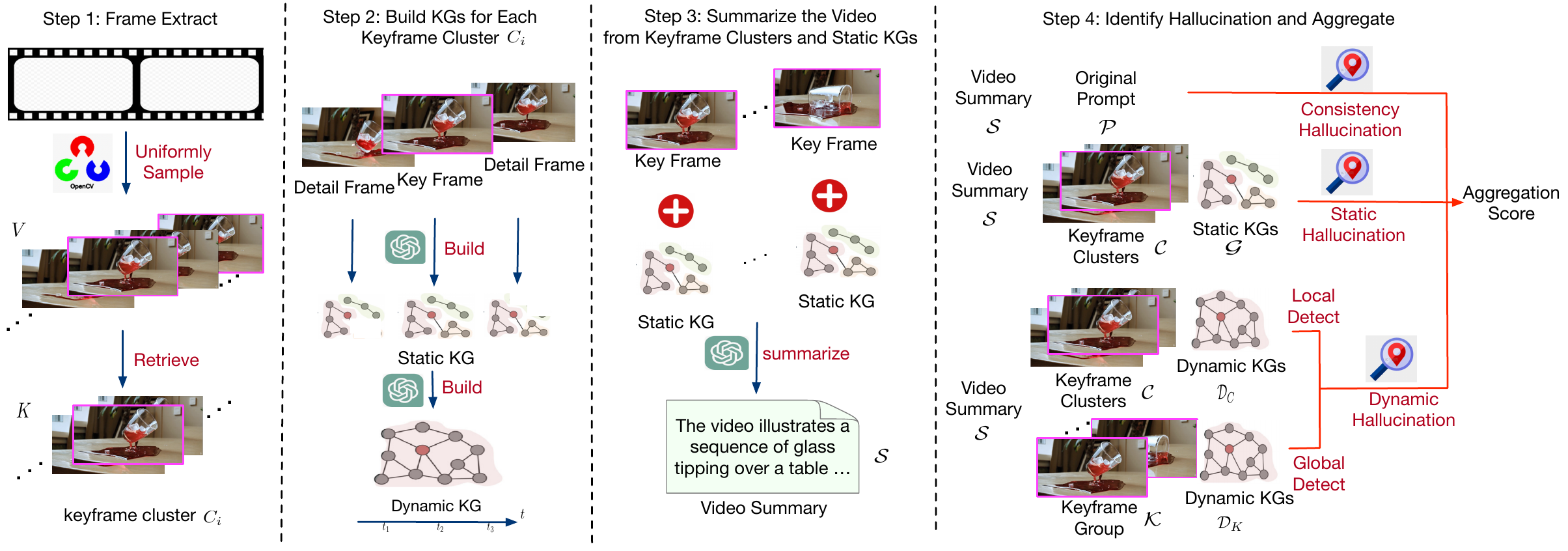}
    
  \caption{Unified hallucination detection for T2V generations.}\label{fig:sora_framework}
  
\end{figure*}

\section{Hallucinations}
T2V generation models often suffer from various types of hallucinations, including prompt consistency hallucinations, static hallucinations, and dynamic hallucinations, which pose significant challenges in generating realistic and coherent videos. Prompt consistency hallucinations occur when the generated video fails to accurately represent the content described in the input prompt, resulting in a mismatch between the textual description and the visual output. This type of hallucination arises due to the model's inability to fully comprehend the semantic meaning of the input prompt and its failure to translate the textual information into consistent visual representations. Static hallucinations arise due to the limitations of the model in understanding and representing the spatial relationships, physical properties, and semantic consistency of objects and scenes within individual frames. On the other hand, dynamic hallucinations occur because of the model's inability to capture the temporal coherence, realistic motion, and plausible interactions of objects across video frames. In this paper, we provide a formal definition for each type of hallucination and present examples of static and dynamic hallucinations.

\subsection{Hallucination Categories}

\begin{definition}[Prompt Consistency Hallucination (PCH)]
Given an input prompt $P$ and the corresponding generated video $V$, PCH occurs when the content of $V$ fails to accurately represent the semantic information conveyed in $P$, resulting in a discrepancy between the textual description and the visual output.
\end{definition}

\begin{definition}[Static Hallucination (SH)]
Static hallucinations refer to the presence of unrealistic, inconsistent, or contextually inappropriate objects, textures, or scenes within individual frames of the generated video $V$. These hallucinations can be formally defined as the existence of elements $E$ within a frame $f_i$ of $V$ that deviate from the expected visual coherence and realism based on the input prompt $P$ and the overall context of the video.
\end{definition}

\begin{definition}[Dynamic Hallucination (DH)]
Dynamic hallucinations encompass temporal inconsistencies and abnormalities in the motion and behavior of objects or entities across the frames of the generated video $V$. These hallucinations can be formally defined as the presence of unrealistic, erratic, or discontinuous movements or changes of elements from consecutive frame $f_i$ to $f_{i+j}$ of $V$, which deviate from the natural expectations and temporal coherence based on the input prompt $P$ and the general principles of object motion and behavior.
\end{definition}

\subsection{Static Hallucination Examples}
\begin{enumerate}
\item Geometric structure irrationality: Objects' shapes, proportions, and topologies are inconsistent with real-world common sense, showing irrational deformations, omissions, redundancies, discontinuities, or inconsistencies.
\item Biological structure irrationality: Biological organs' structures defy real-world common sense, presenting illogical deformations, omissions, redundancies, discontinuities, or inconsistencies.
\item Lighting, shadow, and material physical inaccuracy: The direction, intensity, color of lighting, shape of shadows, and surface material properties in images contradict physical laws and don't match the objects' position, environmental conditions, or the physical properties of materials in the real world.
\item Color distribution disharmony: Color distribution and combinations in images don't follow natural statistical laws, showing unrealistic, disharmonious, or abnormal color combinations.
\item Depth of field and focal length unreality: The depth of field and perspective relationships in images defy physical laws and photography principles, contradicting the spatial positioning of objects.
\item Object composition and scene semantic inconsistency: The arrangement of objects and scene settings in images violate common sense logic, semantic constraints, and scenario-specific norms, featuring irrational, contradictory, rare, or impossible combinations and setups.
\item Motion and blur inconsistency: The direction, extent, and trajectory of motion blur in images contradict the motion state and speed of moving objects, presenting irrational or contradictory motion blur effects.
\item Physical phenomenon inauthenticity: Various physical phenomena in images, like reflection and refraction, defy physical laws, conflicting with the materials, shapes, positions, and environmental conditions of objects.
\item Image quality inconsistency: The overall and local quality of images, like resolution, is inconsistent, showing noticeable quality differences or traces of post-processing.
\end{enumerate}
\subsection{Dynamic Hallucination Examples}
\begin{enumerate}
\item Clipping: Unnatural overlapping and intersection that the boundaries of object models do not correspond to their actual physical relationships.
\item Implausible fusion: Implausible/Unnatural fusion between objects, such as two objects gradually becoming one object when interacting.
\item Implausible appearance or disappearance: The sudden appearance or disappearance of an object without reasonable physical interaction.
\item Implausible motion: The unnatural movement of an object, such as movement without support.
\item Implausible transform: Objects undergo unnatural deformation, such as when a solid suddenly turns into a fluid.
\item Implausible penetration: The unnatural penetration of an object through another object in interaction.
\item Physical interaction errors: The interaction that should have occurred did not occur/An interaction that should not have occurred occurred.
\item Logical interaction error: Timing error/Timing reversal.
\item Other hallucination: All other implausible phenomena except those mentioned above that violate the physical laws and logic of the real world.
\end{enumerate}

We have formally defined three main categories of hallucinations in T2V generation: prompt consistency hallucinations, static hallucinations, and dynamic hallucinations. We have also provided detailed examples of static and dynamic hallucinations, highlighting the various ways in which generated videos can deviate from real-world expectations and physical laws. Understanding and categorizing these hallucinations is crucial for developing effective methods to identify and mitigate them in T2V models.

\section{Sora Detector}

As shown in Figure \ref{fig:sora_framework}, we present Sora Detector, a comprehensive methodology for detecting hallucination problems in videos generated by large T2V models, such as Sora developed by OpenAI. The proposed approach aims to identify inconsistencies, anomalies, and hallucinations in the generated video content by leveraging keyframe extraction, object and relationship detection, knowledge graph construction, and multimodal large language models. The motivation behind developing Sora Detector is to ensure the reliability and trustworthiness of videos generated by large T2V models, as hallucinations can lead to misinformation and confusion.

To validate the hallucination detection process, we first check the premise of the original video description (prompt) used for generating the video. This step is crucial because if the prompt $P$ is inconsistent with real-world physics and logic, the generated video $V$ is likely to contain inherent hallucinations that are not necessarily a result of the model's limitations. We define the set of all possible prompts as $\mathcal{P}$ and the set of all possible generated videos as $\mathcal{V}$. The video generation process can be represented as a function $f: \mathcal{P} \rightarrow \mathcal{V}$. If $P \notin \mathcal{P}_{\text{valid}}$, where $\mathcal{P}_{\text{valid}}$ is the set of prompts that adhere to real-world constraints, then detecting hallucinations in $V = f(P)$ would not provide meaningful insights into the model's performance. Therefore, we focus on videos generated from prompts $P \in \mathcal{P}_{\text{valid}}$, enabling us to assess the model's ability to generate realistic and consistent content.

\subsection{Frame Extraction}

The first step of Sora Detector is keyframe extraction~\citep{guan2012keypoint}. Let $V = \{f_1, f_2, \ldots, f_n\}$ be $n$ uniform frames of a video extracted by the OpenCV library~\citep{OpenCV}. Keyframes are representative frames $K \subseteq V$ that summarize significant moments or events in the video. Keyframe extraction helps to reduce the computational complexity of the analysis by focusing on the most informative frames rather than processing every single frame in the video. The set of all keyframes extracted from the video is defined as the keyframe group $K = \{k_1, k_2, \ldots, k_m\}$, where $m$ is the number of keyframes. The extraction of keyframes serves three primary purposes: (1) summarizing the video content for consistency checking with the original prompt, (2) supporting static hallucination detection, and (3) assisting in subsequent dynamic hallucination detection. In addition, to capture more detailed information, we enrich each keyframe $k_i \in K$ by sampling its surrounding frames, referred to as $d$-th detail frame $k^d_i$ for keyframe $k_i$. Each keyframe $k_i \in K$ and its surrounding $m_i$ detail frames are referred to as a keyframe cluster $C_i = \{k_i, \{k^d_i\}_{d=1}^{m_i} \}$, where $m_i$ is the number of detail frames in the $i$-th keyframe cluster. All keyframe clusters $C = \{C_i\}_{i=1}^m = \Bigl\{k_i, \{k^d_i\}_{d=1}^{m_i}\Bigl\}_{i=1}^m $ provide additional context and help in detecting subtle inconsistencies that may not be apparent in the keyframes alone.

We propose a video keyframe extraction method that operates in three steps:
(1) perform uniform sampling of video frames; (2) utilize neural networks to capture the visual features of these uniformly sampled frames and apply a clustering approach based on density peaks to identify and retain the most representative frames—the top $m$ frames—as keyframes; (3) extract additional detail frames adjacent to each keyframe based on predefined criteria to enhance the detection accuracy of dynamic hallucination detection.

We employ the OpenCV library to carry out the uniform sampling process, retrieving one frame every 5 frames. At the heart of the employed density peak clustering algorithm is the computation of each frame's local density $\rho$ and relative distance $\delta$. The local density $\rho$ is calculated to assess the concentration of frames within a specified radius in the feature space around a frame. This is computed as follows:
\begin{eqnarray}\label{eqnarray1}
		\rho_{i}=\sum K\left(d_{(f_{i}, f_{j})}\right),
\end{eqnarray}
where $K(d(f_{i}, f_{j}))$ denotes a kernel function that gauges the impact of the distance $d(f_{i}, f_{j})$ separating frames $f_{i}$ and $f_{j}$. Typical kernel functions include but are not limited to the cutoff kernel and Gaussian kernel. The local density using the cutoff kernel function is determined as follows:
\begin{eqnarray}\label{eqnarray2}
\rho_{i}=\sum_{j \in\left\{V-\{f_{i}\}\right\}} \chi\left(d_{i j}-d_{c}\right),
\end{eqnarray}
\begin{eqnarray}\label{eqnarray2.1}
\chi(x)=\left\{\begin{array}{ll}
1, & x<0 \\
0, & x \geq 0
\end{array}\right.
\end{eqnarray}
where $d_{i j}=d(f_{i}, f_{j})$, and $d_{c}$ is a cutoff distance. Alternatively, utilizing the Gaussian kernel function as follows
\begin{eqnarray}\label{eqnarray3}
\rho_{i}=\sum_{j \in\left\{V-\{f_{i}\}\right\}} e^{-\left(\frac{d_{i j}}{d_{c}}\right)^{2}}.
\end{eqnarray}
Furthermore, the relative distance $\delta$ represents the distance between a frame and those frames with a higher density than itself.
\begin{eqnarray}\label{eqnarray3}
\delta_{i}=\left\{\begin{array}{ll}
\min _{j \in V^{i}}\left\{d_{i j}\right\}, & V^{i} \neq \varnothing \\
\max _{j \in V}\left\{d_{i j}\right\}, & V^{i}=\varnothing
\end{array}\right.
\end{eqnarray}
\begin{eqnarray}\label{eqnarray3}
V^{i}=\left\{k \in V: \rho_{k}>\rho_{i}\right\}.
\end{eqnarray}
For frames with non-maximum local density, locate all frames that have a higher local density than frame $f_{i}$, and among these frames, identify the frame $f_{j}$ that is closest to frame $f_{i}$; then the relative distance $d_{i,j}$ between frame $f_{i}$ and frame $f_{j}$ is $\delta_{i}$. As for the frame with maximum local density, its relative distance is the maximum distance among all other frames. Eventually, the algorithm posits that frames exhibiting concurrently high values of local density and relative distance are poised to serve as more representative and distinctive keyframes. Consequently, utilizing the criterion $\gamma=\rho \cdot \delta$, frames with the largest $\gamma$ values are selected as the ultimate keyframes.

Following the extraction of keyframes, we propose a detailed frame extraction scheme to capture more intricate information. The fundamental idea is assessing the presence of significant information omitted between successive keyframes, $k_{i}$ and $k_{i+1}$. Detail frames are selected by retaining frames where the difference between successive keyframe pairs exceeds a predetermined threshold $\tau$. This approach balances considerations of compression efficiency and recall accuracy. See the pseudocode for the detail frame extraction in Algorithm \ref{algo:detail_frame}.

\begin{algorithm}
    \renewcommand{\algorithmicrequire}{\textbf{Input:}}
    \renewcommand{\algorithmicensure}{\textbf{Output:}}
  \setstretch{1.2}
  \caption{Detail Frame Extraction}
  \label{algo:detail_frame}
  \begin{algorithmic}[1]
  \Require $V$: Uniform frame group $\{f_1, f_2, \ldots, f_n\}$, $H$: The ordered sets including $1$, the index of keyframes $K$, and $n$, $\mathbf{\tau}$: Decision threshold for adjacent frame similarity.
  \Ensure Detail frame sets $D$

    \For{$k = 0$ \textbf{to} $\text{len}(H)-2$}
        \State $i \gets H[k]$
        \State $j \gets H[k+1]$
    
        \State $\alpha \leftarrow \operatorname{Similar\_function}\left(f_{i}, f_{j}\right)$

        \If{$\alpha > \mathbf{\tau}$}  
            \State $t \leftarrow i$ 
            \For{$d=i+1$ to $j-1$}
                \State $\beta \leftarrow \operatorname{Similar\_function}\left(f_{t}, f_{d}\right)$

                \If{$\beta > \tau$}

                \State $t \leftarrow d$ 
                \State  $D.add \left( f_{d}\right)$

                \EndIf
            \EndFor
        \EndIf
    \EndFor
    \State \textbf{return} $D$
    \end{algorithmic}
\end{algorithm}

\subsection{Knowledge Graph}

Once the keyframes and detail frames are obtained, we employ object detection and interaction recognition techniques to extract object information and their relationships within each frame. This step is crucial for constructing a static knowledge graph that represents the scene at each frame. The static knowledge graph captures the spatial relationships and interactions between objects in the frame, providing a structured representation of the video content. These tasks can be achieved by several separate AI models, such as object detection, interaction recognition, and knowledge graph construction, or by a multimodal large language model like GPT-4. 

\subsubsection{Construction of Knowledge Graph}
To construct the static knowledge graph~\citep{Pan_2024}, we leverage the capabilities of GPT-4, to detect objects, recognize their relationships, and generate triples based on the frames. GPT-4 has the ability to understand and generate natural language descriptions of visual content and output this knowledge in the form of triples. By utilizing the vision-language understanding capabilities of GPT-4, we can effectively extract the semantic information from the video frames and represent it in a structured manner. The process of constructing the static knowledge graph using GPT-4 involves several steps:

\begin{enumerate}
\item  Object Detection: For all keyframes and the surrounding detail frames $c_i \in C$, GPT-4 analyzes the visual content to detect and identify the objects present in the scene. The set of detected objects in frame $c_i$ is denoted as $O_i = \{o_1, o_2, \ldots, o_{n_{o_i}}\}$, where $n_{o_i}$ is the number of objects in frame $c_i$.

\item Relationship Recognition: Once the objects are detected, GPT-4 determines the spatial relationships and interactions between the detected objects in each frame $c_i$. Let $R_i = \{r_1, r_2, \ldots, r_{n_{r_i}}\}$ be the set of relationships identified in frame $c_i$, where $n_{r_i}$ is the number of relationships in frame $c_i$. 

\item Triple Generation: Based on the detected objects $O_i$ and their relationships $R_i$, GPT-4 generates triples in the form of (subject, predicate, object) to represent the identified relationships. The set of generated triples for frame $c_i$ is denoted as $T_i = \{t_1, t_2, \ldots, t_{n_{t_i}}\}$, where $n_{t_i}$ is the number of triples in frame $c_i$. Each triple is represented as $T_i = (O_i, R_i, O_i)$. For example, if a frame depicts a person sitting on a chair, GPT-4 would generate a triple such as (person, sitting\textunderscore on, chair).

\end{enumerate}

The static knowledge graph for frame $c_i$ is then constructed using the generated triples $T_i$, where the objects $O_i$ serve as the nodes and the relationships $R_i$ serve as the edges. The static knowledge graph for frame $c_i$ is denoted as $G_i = (V_i, E_i)$, where $V_i = O_i$ is the set of nodes (objects) and $E_i$ is the set of edges with $R_i$ being the set of all possible relationships in frame $c_i$. By applying this process to each frame $c_i \in C$, we obtain a set of static knowledge graphs $\mathcal{G} = \{G_1, G_2, \ldots, G_{nc}\}$ that capture the semantic information and relationships within all keyframe clusters including all keyframes and surrounding detail frames.

\subsubsection{Advantages of Static Knowledge Graph }
The generated triples form the nodes and edges of the static knowledge graph, providing a comprehensive representation of the scene. Constructing the static knowledge graph using GPT-4 offers several advantages over direct hallucination detection based on images:

\textbf{1. Semantic Understanding:} It can capture the semantic meaning of the relationships between objects, going beyond simple spatial relationships. It can infer higher-level concepts and actions based on the context of the scene. For instance, if a frame shows a person holding a book, GPT-4 can generate a triple-like (person, reading, book), indicating the action of reading. This rich semantic representation enhances the expressiveness and interpretability of the static knowledge graph~\citep{ji_survey_kg_2022}, enabling a deeper understanding of the video content.

\textbf{2. Handling Complexity:} It can handle ambiguous or complex scenes by generating multiple triples that capture different aspects of the relationships between objects. In scenes with multiple objects and interactions, GPT-4 can generate triples for each relevant relationship, providing a comprehensive representation of the scene. This ability to handle complexity is particularly valuable in videos with diverse and dynamic content.

\textbf{3. Temporal Evolution:}
Furthermore, the static knowledge graph can evolve into a dynamic knowledge graph by incorporating the temporal dimension~\citep{cai2024survey}. By analyzing the evolution of the knowledge graph over time, we can identify inconsistencies in object interactions and detect anomalies that may not be apparent in individual frames. This temporal analysis allows for the detection of dynamic hallucinations, where objects or relationships change in an inconsistent or implausible manner over time.

The construction of the static knowledge graph using GPT-4 is a critical step in the Sora Detector methodology. It provides a structured and semantic representation of the video content, capturing the relationships and interactions between objects in a way that goes beyond simple visual features. The knowledge graph serves as a common representation that can be used across different sub-tasks of hallucination detection, such as consistency checking with the original video description, static hallucination detection within individual frames, and dynamic hallucination detection across frames.

\subsection{Consistency Hallucination Detection}

Consistency hallucination detection is a crucial step in the Sora Detector methodology that aims to identify discrepancies between the generated video content and the original video generation prompt. This step is performed when the original prompt $P$ is available, providing a reference for the intended content and narrative of the video.

Let $S$ be the video content summary obtained from a multimodal large language model by analyzing all frames in the keyframe clusters $C = \{C_1, C_2, \ldots, C_m\}$ and the corresponding static knowledge graphs $\mathcal{G} = \{G_1, G_2, \ldots, G_{nc}\}$. The video content summary $S$ encapsulates the main events, objects, and interactions present in the video, providing a concise representation of the video's content.

To perform consistency hallucination detection, we define a similarity function $\mathcal{H}_c: \mathcal{P} \times \mathcal{S} \rightarrow [0, 1]$, where $\mathcal{P}$ is the set of all possible prompts and $\mathcal{S}$ is the set of all possible video content summaries. The function $\mathcal{H}_c(P, S)$ measures the similarity between the original prompt $P$ and the video content summary $S$. A higher value of $\mathcal{H}_c(P, S)$ indicates a greater consistency between the prompt and the generated video, while a lower value suggests the presence of consistency hallucinations.

Let $\tau$ be a predefined threshold for consistency hallucination detection. If $\mathcal{H}_c(P, S) < \tau$, we consider the generated video to have consistency hallucinations, as the video content summary $S$ deviates significantly from the intended content specified in the prompt $P$.

For example, let $P_1$ be a prompt specifying a scene where a person is walking in a park, and let $S_1$ be the video content summary of the generated video. If $S_1$ reveals that the generated video shows the person in an entirely different setting, such as a busy city street, the similarity function $\mathcal{H}_c(P_1, S_1)$ would yield a low value, indicating a consistency hallucination. Similarly, if $P_2$ describes a specific sequence of events, but the video content summary $S_2$ indicates a different order or omits certain events altogether, $\mathcal{H}_c(P_2, S_2)$ would also result in a low value, flagging it as a consistency hallucination.

\subsubsection{Motivation of Consistency Hallucination Detection}

The consistency hallucination detection step serves as a high-level verification of the generated video's adherence to the original prompt. It helps to identify cases where the video generation process has strayed from the intended content, resulting in videos that do not align with the desired narrative or context. If the video content summary is found to be consistent with the original prompt, the prompt is considered as the video summary and is used to assist in the subsequent static and dynamic hallucination detection steps. 

In the static hallucination detection step, the video summary is used to provide additional context and expectations for the visual content of each frame. By comparing the static knowledge graph and the visual content of each frame with the video summary, we can identify inconsistencies and anomalies that deviate from the expected content. This helps to filter out false positives and focus on anomalies that are relevant to the overall context of the video~\citep{saligrama2010video}.

Similarly, in the dynamic hallucination detection step, the video summary is used to identify temporal inconsistencies and anomalies across video frames. By comparing the dynamic knowledge graph and the temporal evolution of the video with the video summary, we can identify deviations from the expected storyline or context. This helps to detect anomalies that may not be apparent from the dynamic knowledge graph alone, enhancing the accuracy and effectiveness of the dynamic hallucination detection process.

Furthermore, the consistency hallucination detection step can provide valuable insights into the limitations and challenges of the video generation process. By analyzing the discrepancies between the generated video and the original prompt, we can identify patterns and trends in the types of consistency hallucinations that occur. This information can be used to improve video generation models and algorithms, enabling the creation of more accurate and coherent videos.

\subsection{Static Hallucination Detection}

Static hallucination detection aims to identify anomalies and hallucinations within individual keyframes and detail frames, i.e., keyframe clusters, denoted as $C = \{C_1, C_2, \ldots, C_m\}$. Let $c_i$ represent the $i$-th frame in all keyframe clusters $C$. The static knowledge graph for frame $c_i$ is denoted as $G_{i} = (V_{i}, E_{i})$. To enhance the effectiveness of static hallucination detection, we also leverage the video content summary $S$ obtained from the consistency hallucination detection step. By comparing the static knowledge graph $G_{i}$ and the visual content of each frame $c_{i}$ with the video content summary $S$, we can identify inconsistencies and anomalies that deviate from the expected storyline or context.

We define a static hallucination detection function $\mathcal{H}_s: \mathcal{C} \times \mathcal{G} \times \mathcal{S} \rightarrow [0, 10]$, where $\mathcal{C}$ is the set of all keyframe clusters, $\mathcal{G}$ is the set of all static knowledge graphs, and $\mathcal{S}$ is the set of all video content summaries. The function $\mathcal{H}_s(c_{i}, G_{i}, S)$ measures the degree of static hallucination in frame $c_{i}$ by comparing it with the corresponding static knowledge graph $G_{i}$ and the video content summary $S$. A higher value of $\mathcal{H}_s(c_{i}, G_{i}, S)$ indicates a more severe static hallucination. To compute $\mathcal{H}_s(c_{i}, G_{i}, S)$, we leverage multimodal large language models, such as GPT-4. GPT-4 takes the frame $c_{i}$, the static knowledge graph $G_{i}$, and the video content summary $S$ as inputs and outputs a value in the range $[0, 10]$, along with a description of the detected static hallucination, if any. 

For example, if the video content summary $S$ mentions a person sitting on a chair in a room, but the static knowledge graph $G_{i}$ for frame $c_{i}$ indicates that the person is floating in mid-air, GPT-4 would output a high value of $h_{i}$ and a description highlighting the inconsistency between the expected and observed relationships. Moreover, the video content summary can help in detecting inconsistencies that may not be apparent from the static knowledge graph alone. For instance, if the video content summary mentions a specific location or time period, but the visual content of a frame contains anachronistic or out-of-place objects, this inconsistency can be detected by comparing the frame with the video content summary.

In summary, static hallucination detection plays a crucial role in identifying anomalies and inconsistencies within individual frames of the generated video. By leveraging the static knowledge graph and employing multimodal large language models like GPT-4, we can detect a wide range of hallucinations, including lighting errors, texture anomalies, object shape anomalies, biological form anomalies, and so on.

\subsection{Dynamic Hallucination Detection}

In contrast to static image hallucination detection, video hallucination detection requires considering the consistency and continuity in the temporal dimension. Static hallucination detection primarily focuses on anomalies and inconsistencies within single-frame images, such as lighting errors, texture anomalies, and object shape abnormalities. However, in videos, beyond static hallucinations, the most crucial aspect is dynamic hallucinations in the temporal dimension~\citep{sun2024temporal,chu2020learning}. Videos are composed of a series of consecutive frames, necessitating the consideration of temporal relationships and changes between frames. Ignoring the temporal dimension may lead to missing many hallucination issues that cannot be detected in single-frame images alone. Dynamic hallucination detection aims to identify temporal inconsistencies, anomalies, and hallucinations across video frames. Temporal hallucinations can manifest in various forms, such as objects appearing, disappearing, or transforming in an inconsistent manner, or characters exhibiting impossible or erratic movements. For example, a temporal hallucination could involve a person suddenly disappearing from one frame to the next without any logical explanation, or an object changing its shape or color abruptly. These anomalies may be difficult to perceive in single-frame images but become evident in the temporal dimension of the video~\citep{chang2022video}.

\subsubsection{The Construction of Dynamic Knowledge Graph}

To detect these dynamic hallucinations, we consider each keyframe and its surrounding detail frames as a keyframe cluster and perform temporal dynamic hallucination detection. Based on the static knowledge graph for each frame, for each keyframe cluster, we construct a dynamic knowledge graph $D_{C_i}$ representing the changes in object interactions within keyframe cluster $C_i$. The construction of the dynamic knowledge graph $D_{C_i}$ involves the following steps:

\begin{enumerate}
    \item Object Tracking: We track the objects detected in the static knowledge graphs $G_{i,j} = (V_{i,j}, E_{i,j})$ for frame $c_{i,j}$ within the keyframe cluster $C_i$. 

    \item Temporal Relationships: For each pair of consecutive frames $(c_{i,j}, c_{i,j+1})$ in the keyframe cluster $C_i$, we analyze the changes in object relationships and interactions. Let $R_{i,(j,j+1)}^d$ be the set of dynamic relationships between objects in frames $c_{i,j}$ and $c_{i,j+1}$. These temporal relationships capture the changes in object positions, interactions, and attributes between consecutive frames.

    \item Dynamic Knowledge Graph Construction: The dynamic knowledge graph $D_{C_i}$ is constructed using the tracked objects $O_{C_i}$ as nodes and the temporal relationships $R_{i,(j,j+1)}^d$ as edges. 
\end{enumerate}

In addition to constructing dynamic knowledge graphs~\citep{trivedi2017know} for each keyframe cluster, we also construct a dynamic knowledge graph $D_K$ for the entire keyframe group $K$. The construction of $D_K$ follows a similar process as described above, but it considers the keyframes $k_i \in K$ instead of the individual frames within a keyframe cluster. The objects $O_{K}$ and temporal relationships $R_{K}^d$ are tracked and identified across the keyframes in $K$, resulting in a higher-level representation of the changes in object interactions throughout the entire video. The dynamic knowledge graphs $D_{C_i}$ and $D_K$ capture the temporal evolution of object relationships and interactions within keyframe clusters and across the entire video, respectively.

\subsubsection{The Procedure of Dynamic Hallucination Detection}

Dynamic hallucination detection is divided into two parts: local detection and global detection. Local detection focuses on identifying hallucinations within each keyframe cluster $C_i$, while global detection identifies hallucinations across all keyframes $K$.

\paragraph{Local Detection.} This allows us to detect hallucinations that occur within short time intervals and are specific to a particular scene or event. For instance, within a keyframe cluster, an object might suddenly disappear for a few frames and then reappear, which can be identified through local detection. Let $\mathcal{H}_d^l: \mathcal{C} \times \mathcal{D} \times \mathcal{S} \rightarrow [0, 10]$ be the local dynamic hallucination detection function, where $\mathcal{D}$ is the set of all dynamic knowledge graphs and $\mathcal{S}$ is the set of all video content summaries. The function $\mathcal{H}_d^l(C_i, D_{C_i}, S)$ measures the degree of local dynamic hallucination in keyframe cluster $C_i$ by comparing it with the corresponding dynamic knowledge graph $D_{C_i}$ for keyframe cluster $C_i$ and the video content summary $S$. A higher value of $\mathcal{H}_d^l(C_i, D_{C_i}, S)$ indicates a more severe local dynamic hallucination.

\paragraph{Global Detection.} This identifies hallucinations in the entire keyframe group, considering the full length of the video. Global detection helps to identify hallucinations that persist throughout the video or involve inconsistencies between different scenes or events. For example, a character's clothing color may change between different scenes, or an object may appear in different shapes at different points in the video. These anomalies that span across multiple keyframe clusters require global detection to be identified. Let $\mathcal{H}_d^g: \mathcal{K} \times \mathcal{D} \times \mathcal{S} \rightarrow [0, 10]$ be the global dynamic hallucination detection function, where $\mathcal{K}$ is the set of all keyframes. The function $\mathcal{H}_d^g(K, D_K, S)$ measures the degree of global dynamic hallucination across all keyframes $K$ by comparing them with the set of dynamic knowledge graphs $D_K$ for the keyframe group and the video content summary $S$. A higher value of $\mathcal{H}_d^g(K, D_K, S)$ indicates a more severe global dynamic hallucination.

To compute $\mathcal{H}_d^l(C_i, D_{C_i}, S)$ and $\mathcal{H}_d^g(K, D_K, S)$, we leverage multimodal large language models, such as GPT-4. GPT-4 takes the keyframe cluster $C_i$, the dynamic knowledge graph $D_i$, and the video content summary $S$ as inputs for local detection, and the set of keyframes $K$, the set of dynamic knowledge graphs $\mathcal{D}$, and the video content summary $S$ as inputs for global detection. It outputs a value in the range $[0, 10]$, along with a description of the detected dynamic hallucination, if any.

\subsubsection{The Connection between Static and Dynamic Hallucination Detection}

Moreover, the results from static hallucination detection provide valuable information about the anomalies and inconsistencies present in individual frames. By incorporating these results into the dynamic hallucination detection process, we can prioritize the analysis of frames that have already been identified as containing static hallucinations. This targeted approach reduces computational overhead and improves the efficiency of dynamic hallucination detection~\citep{zhou2020noh}.

For instance, if a frame has been flagged as containing a texture anomaly during static hallucination detection, we can focus on analyzing the temporal consistency of that specific texture anomaly across the keyframe cluster. This allows us to determine whether the anomaly is a localized issue or persists throughout the video, providing a more comprehensive understanding of the hallucination.

Furthermore, the static hallucination detection results can serve as a starting point for identifying dynamic hallucinations. If multiple frames within a keyframe cluster are flagged as containing similar static hallucinations, such as object shape anomalies, we can investigate the temporal evolution of those anomalies~\citep{wang2015localizing} to determine if they are consistent across the cluster or if they exhibit inconsistent or erratic behavior, indicating a dynamic hallucination.

In summary, dynamic hallucination detection plays a crucial role in identifying temporal inconsistencies, anomalies, and hallucinations in videos. By considering the continuity and consistency over time, dynamic hallucination detection enables the identification of anomalies that may be challenging to discover in single-frame images. The combination of local and global detection approaches allows Sora Detector to comprehensively identify various hallucination issues in videos. The integration of the video content summary and the results from static hallucination detection further enhances the accuracy and efficiency of the dynamic hallucination detection process, ultimately improving the quality and reliability of the generated video content.

\begin{figure*}[t]
    \centering
    \includegraphics[width=1\textwidth]{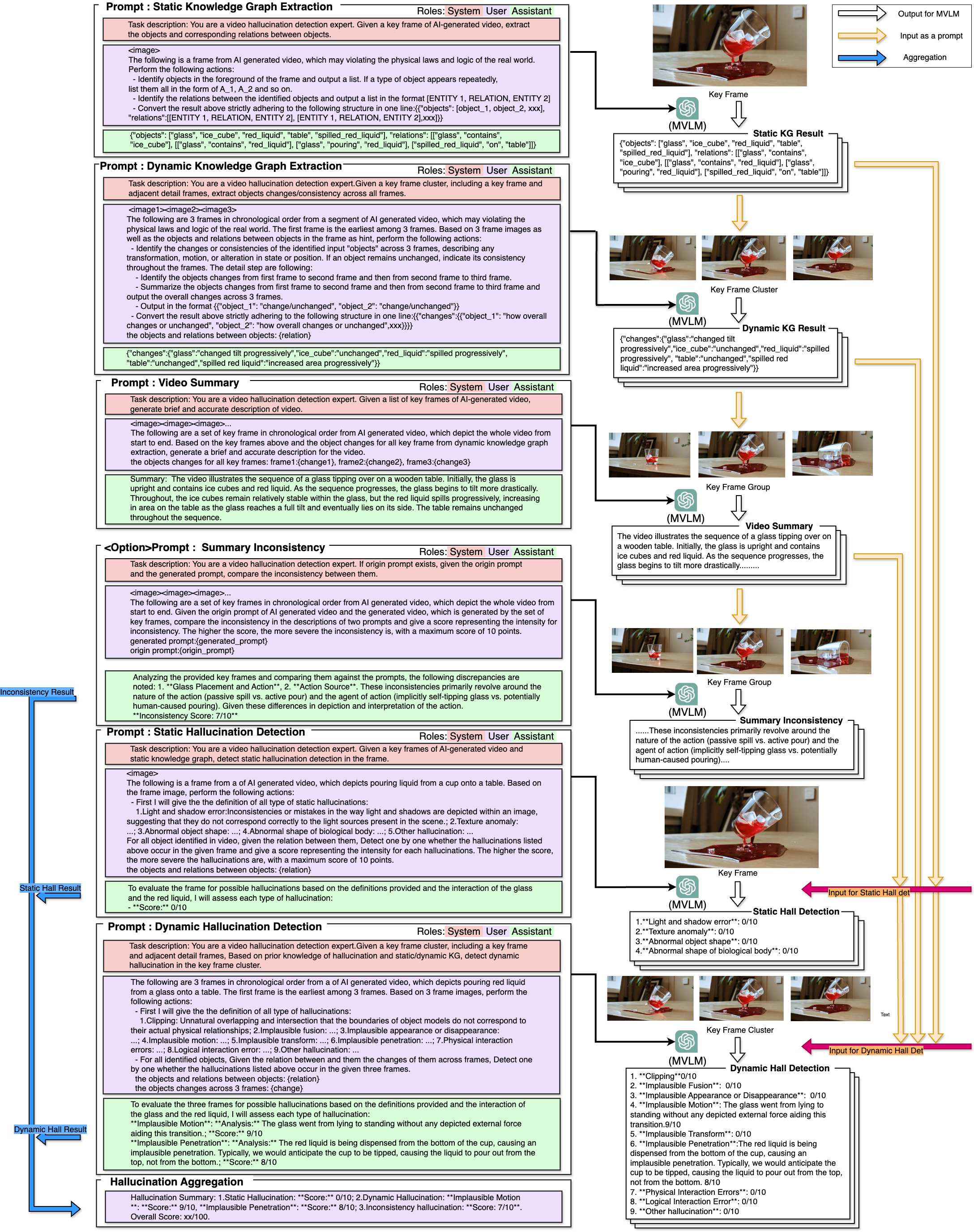}
    
    \caption{The whole procedure and detailed prompt examples for each step of Sora Detector for Sora video ``Liquid Pouring''.}
    
    \label{fig:prompt_example}
\end{figure*}

\subsection{Hallucination Aggregation}

Finally, we aggregate the hallucination problems identified from the three detection steps: consistency hallucination, static hallucination, and dynamic hallucination. Let $\mathcal{H}_c$, $\mathcal{H}_s$, and $\mathcal{H}_d$ denote the sets of consistency, static, and dynamic hallucinations, respectively. We define an aggregation function $\mathcal{A}: \mathcal{H}_c \times \mathcal{H}_s \times \mathcal{H}_d \rightarrow \mathcal{H}$, where $\mathcal{H}$ is the set of all aggregated hallucinations. The function $\mathcal{A}(H_c, H_s, H_d)$ combines the hallucinations from the three detection steps into a comprehensive set of hallucinations $H \in \mathcal{H}$.
Each aggregated hallucination $h \in H$ is associated with a severity score, where a higher value indicates a more severe hallucination. The severity score is computed based on the individual severity scores of the consistency, static, and dynamic hallucinations that contribute to the aggregated hallucination. Let $s_c$, $s_s$, and $s_d$ denote the severity scores of consistency, static, and dynamic hallucinations, respectively. The severity score of an aggregated hallucination $h$ is given by:
\begin{equation}
s_h = f(s_c, s_s, s_d)
\end{equation}
where $f$ is a function that combines the individual severity scores, such as a weighted average or maximum value. The aggregated results provide insights into the types and severity of hallucinations, as well as their temporal and spatial distribution throughout the video.

\subsection{Hallucination Detection Agent}

The proposed Sora Detector methodology offers a systematic approach to identifying and analyzing hallucination problems in videos generated by multimodal large language models. To automate the hallucination detection process for each input video, we have developed the Sora Detector Agent based on LLM~\citep{wu2023autogen,chu2024professional,guan2023intelligent}, an intelligent system capable of executing all the aforementioned steps and generating a comprehensive video quality analysis report. The Sora Detector Agent seamlessly integrates the various components of the methodology, ensuring a streamlined and efficient analysis of the input video.

The video quality analysis report generated by the Sora Detector Agent consists of two main sections: a video quality summary and a detailed hallucination analysis. The video quality summary provides an overall assessment of the generated video, highlighting the presence and severity of hallucinations across the three detection dimensions: consistency, static, and dynamic. This summary offers a high-level overview of the video's quality and reliability, enabling users to quickly grasp the extent of hallucination issues within the video. In addition to the summary, the report includes a meticulous breakdown of each step in the Sora Detector methodology. The detailed hallucination analysis section presents the results of consistency, static, and dynamic hallucination detection.

\begin{figure*}[t]
    \centering
    \includegraphics[width=1\textwidth]{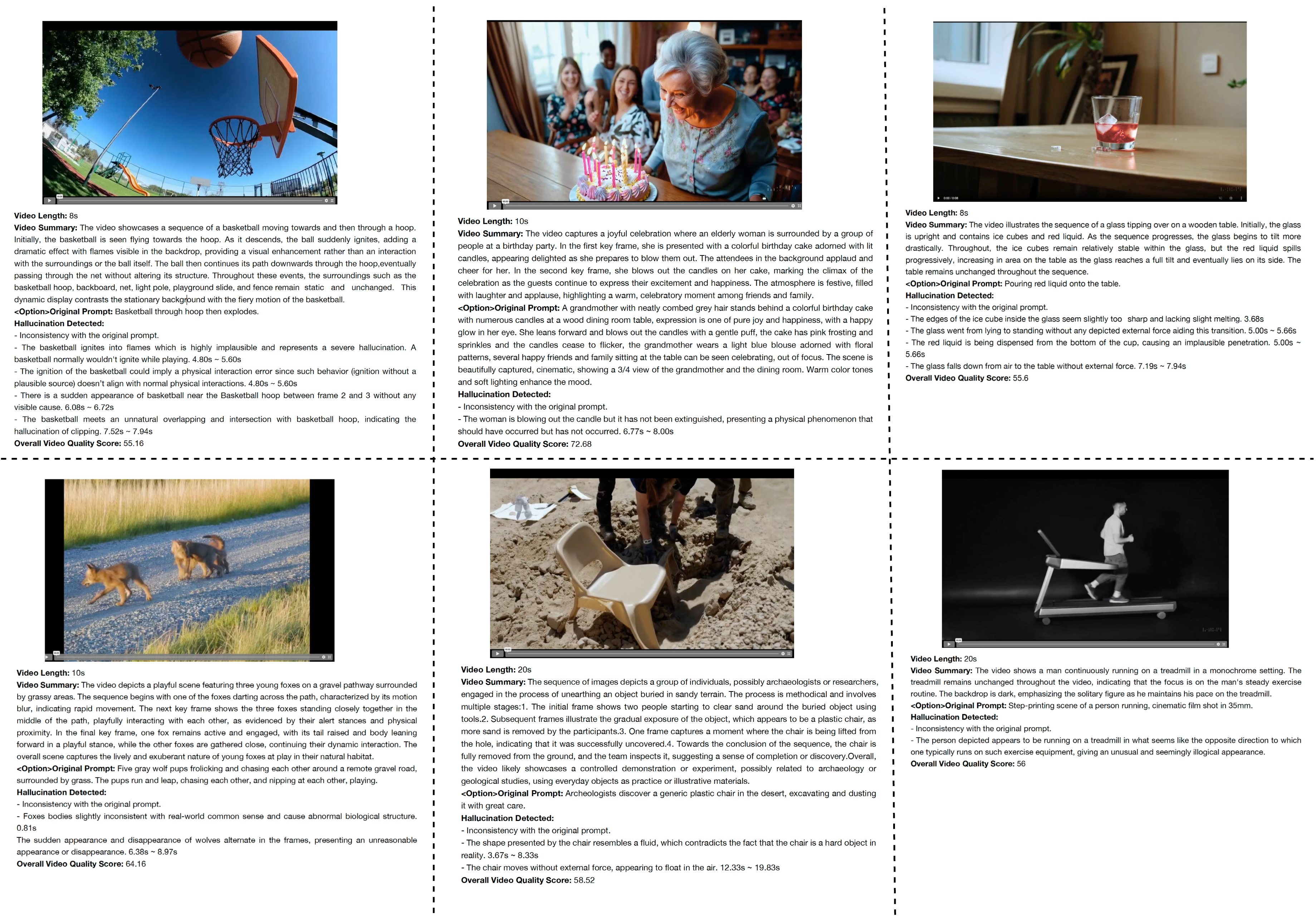}
    
    \caption{Video quality analysis summaries of 6 Sora hallucination videos. }
    
    \label{fig:video_summary}
\end{figure*}

\begin{figure*}[t]
    \centering
    \includegraphics[width=1\textwidth]{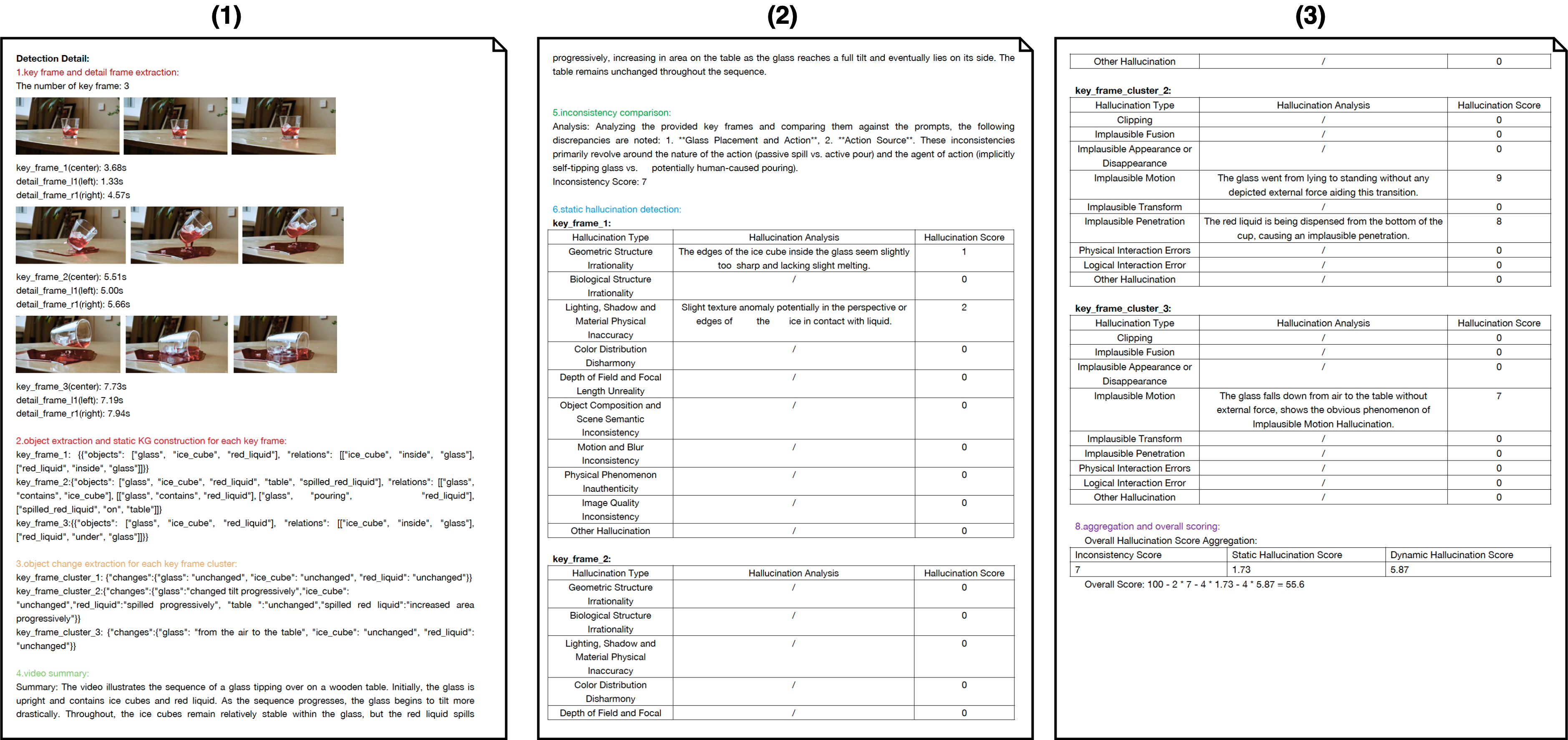}
    
    \caption{Detailed hallucination analysis of video ``Liquid Pouring''.}
    
    \label{fig:video_details}
\end{figure*}

\section{Sora Experiments}
\subsection{Experiment Settings}
\subsubsection{Sora Dataset}
To comprehensively evaluate the effectiveness of our proposed Sora Detector methodology, we have meticulously curated a diverse dataset comprising videos generated by the state-of-the-art Sora model, a groundbreaking T2V generation system developed by OpenAI. Our experimental evaluation primarily focuses on six carefully selected videos from the Sora-generated segments \footnote{https://openai.com/sora}\footnote{https://openai.com/research/video-generation-models-as-world-simulators}, each exhibiting distinct and intriguing hallucination phenomena. These videos encompass a wide spectrum of scenarios and themes, including ``Liquid Pouring'', ``Person Running'', ``Wolves Chasing'', ``Basketball Exploding'', ``Grandmother Blowing Out Candles'' and ``Chair Excavating''. To detect and analyze the video hallucinations, we employ GPT-4V, a highly advanced multimodel large language model, which implements our proposed Sora Detector pipeline for each video. The whole procedure and detailed prompt examples for each step are provided in Figure \ref{fig:prompt_example}.

\subsubsection{Evaluation metrics}
To quantitatively assess the quality of the Sora-generated videos and the severity of the hallucination problem, we have developed a video quality score tailored specifically for video hallucinations. The lower the score, the more severe the hallucination problem, indicating a higher degree of visual anomalies and inconsistencies within the generated video content. The score calculation is defined as follows:
\begin{eqnarray}
\label{eqnarray2}
VideoQualityScore &=& 100 - \alpha  * \mathcal{H}_c - \beta  * \sum_{i=1}^{K}T_i*\sum_{j=1}^{S}(\mathcal{H}_s)_{ij} \nonumber \\
      &-& \gamma * \sum_{i=1}^{K}T_i*\sum_{j=1}^{D}(\mathcal{H}_d)_{ij}, \nonumber
\end{eqnarray}
where $T_i$ represents the proportion of the i-th keyframe to the total video duration, effectively capturing the temporal significance of each frame within the video. $S$ and $D$ denote the number of static hallucinations and dynamic hallucinations, respectively, as defined in the previous sections. $\alpha$, $\beta$, and $\gamma$ are hyperparameters that represent the intensity of each respective hallucination. By default, these hyperparameters are set to values of $2$, $4$, and $4$, respectively, based on empirical observations and extensive experimentation. However, these values can be fine-tuned and adjusted depending on the specific requirements and characteristics of the dataset being analyzed, allowing for a more tailored and adaptive approach to video hallucination assessment.

\subsection{Results and Analysis}

Our method can effectively identify the hallucination problems in $6$ Sora videos. Figure \ref{fig:video_summary} presents the automatically generated video quality summary reports, which provide an overview of the issues detected in each video. For a more in-depth look at the specific hallucination issues found, Figure \ref{fig:video_details} showcases the detailed video quality reports that were automatically produced by our system. These detailed reports offer a comprehensive pipeline of each step of our system, enabling a thorough understanding of the nature and extent of the issues present. The complete reports for all videos are provided in the Appendix.

Despite the promising performance of our proposed method in detecting various hallucination categories, it does not achieve complete consistency with the ground truth labels for each hallucination category defined in the previous sections. To illustrate this limitation, let us consider the ``Liquid Pouring'' video as an example.
Figure \ref{fig:prompt_example} demonstrates that our method fails to detect the ``Physical phenomenon inauthenticity'' static hallucination in the ``Liquid Pouring'' video. This is because the selected keyframe does not contain sufficient information to identify this specific issue. The inability to capture and analyze the relevant details in the keyframe hinders the detection of the physical phenomenon of inauthenticity hallucination.
This shortcoming highlights the need for further improvements in our approach, particularly in the keyframe selection process and hallucination detection techniques. By enhancing the keyframe selection algorithm to prioritize frames that encompass a wider range of relevant information, we can increase the likelihood of detecting subtle hallucinations.

\section{Benchmark Experiments}
\subsection{Construction of Dataset}
To advance the field of T2V generation and facilitate a more rigorous evaluation of hallucination detection methods, we have carefully constructed a novel benchmark dataset, i.e., \textbf{T2VHaluBench}. This benchmark is specifically designed to assess the performance and robustness of various hallucination detection approaches in the context of T2V synthesis. The dataset comprises a diverse collection of $50$ videos generated by Runway-Gen-2 \footnote{https://research.runwayml.com/gen2}, a state-of-the-art commercial T2V model. To introduce a wide range of hallucinatory phenomena, we augmented the original generation prompts using ChatGPT. The prompts were meticulously crafted to elicit different categories of hallucinations, ensuring a comprehensive and challenging evaluation of detection methods. 

The dataset comprises $50$ videos, each with a minimum duration of $8$ seconds to ensure adequate temporal information for analysis. Among these videos, $46$ videos contain hallucinations, while $4$ videos do not. The videos have an average of $3.02$ hallucinations per video, consisting of $1.08$ static hallucinations and $1.94$ dynamic hallucinations. To enhance the benchmark, $9$ additional videos generated by Sora have been included, out of which $7$ videos exhibit hallucinatory content, and $2$ videos do not contain any hallucinations. This expanded dataset offers a diverse range of examples for studying and evaluating the presence of hallucinations in video content.

To establish ground truth labels, we conducted extensive human annotations on the benchmark videos. The annotation process involved categorizing the videos into three main types of hallucinations, i.e., consistency, static, and dynamic. Consistency hallucinations refer to discrepancies between the generated video and the input prompt, while static and dynamic hallucinations correspond to the presence of unrealistic or impossible objects or actions within the video. The annotation guidelines were carefully designed to strike a balance between strictness and practicality. Annotators were instructed to prioritize the most prominent category of hallucination when multiple categories were present in a single video.

\subsection{Evaluation metrics} 
We calculate recall, precision, and F1 metrics for each hallucination category individually, including static hallucination for each category (SH-multiple), and dynamic hallucination for each category (DH-multiple). Besides, we calculate precision for overall hallucinations (OH), prompt consistency hallucination (PCH), overall static hallucination (SH-binary), and overall dynamic hallucination (DH-binary). In addition, we adopt the same video quality score schema for video hallucinations.

\subsection{Cost Estimation}
For one video hallucination detection, The number of GPT-4 calls is positively correlated with the number of extracted keyframes, specifically $4m+2$ calls for one video hallucination detection, where $m$ denotes the number of extracted keyframes. The average cost per call is \$0.08, so that one video hallucination detection costs \$1.28 as there are 3-4 keyframes in general in our constructed benchmark.

%%%%%%%%%%%%%%%%%%%%%%%%%%%%%%%%%%%%%%%%%%%%%%%%%%%%%%%%%%%%%%%%% 
%%%%%%%%%%%%%%%%%%%%%%%%%%%%%%%%%%%%%%%%%%%%%%%%%%%%%%%%%%%%%%%%%

\begin{table*}[h!]
    \centering

    \caption{Experimental results and Ablation studies of our Sora Detector on our constructed dataset.}
    
    \resizebox{1\columnwidth}{!}{
        \begin{tabular}{l|ccc |ccc |c|c|c|c}
        \toprule
         Method
         & \multicolumn{3}{c}{SH-multiple} & \multicolumn{3}{c}{DH-multiple} & \multicolumn{1}{c}{PCH}&\multicolumn{1}{c}{SH-binary}&\multicolumn{1}{c}{DH-binary}&\multicolumn{1}{c}{OH} \\
        
           \cmidrule(lr){2-4}  \cmidrule(lr){5-7} \cmidrule(lr){8-8} \cmidrule(lr){9-9}\cmidrule(lr){10-10}\cmidrule(lr){11-11}

         & P(\%)  & R(\%)& F1  
         & P     & R(\%)  & F1& P(\%)& P(\%)& P(\%)& P(\%) \\
       
        \midrule
          w/o KG & 14.80 & {25.37} & {18.69} & 2.70 & {4.27} & {3.31}  & {68.00} & {60.00}& {36.00}& {68.00} \\
          w/o static KG & 15.88 & {26.16} & {19.79} & 45.88 & {45.51} & {45.78}  & {94.00}& {62.00}& {64.00}& {94.00}  \\
           w/o dynamic KG &31.66 & {62.68} & {42.05} & 8.00 & {4.72} & {5.97} & {74.00} & {88.00}& {40.00}& {88.00} \\
         Sora Detector (ours) & \textbf{32.78} & \textbf{62.92} & \textbf{43.15} & \textbf{46.91} & \textbf{45.48} & \textbf{46.16} & \textbf{98.00}& \textbf{92.00}& \textbf{64.00}& \textbf{98.00} \\

       \bottomrule
    \end{tabular}}
    
    \label{main_results}
    
\end{table*}

\subsection{Evaluation Results}
Table \ref{main_results} presents the evaluation results for T2V hallucination detection. The overall hallucination detection precision, which aggregates the results from each hallucination category, surpasses $98\%$. This high precision value indicates the effectiveness of our proposed method in accurately identifying hallucinations in the generated videos. However, when examining the performance for specific hallucination types, we observe lower F1 scores for both static and dynamic hallucination detection. The F1 score for static hallucination detection stands at $43.15$, while the F1 score for dynamic hallucination detection is $46.16$. These relatively lower F1 scores suggest that there are challenges in detecting certain categories of hallucinations within each type. The discrepancy between the high overall precision and the lower F1 scores for specific hallucination categories may be attributed to the variations in the manifestation of different hallucination types and the definition of each hallucination.

\subsection{Ablation Study}
To better understand our hallucination detection method, we conduct ablation experiments to demonstrate the effectiveness of different parts of our method. For \textbf{w/o KG}, the ablation experiment identifies the hallucination directly without the knowledge graph construction. For \textbf{w/o static KG}, the ablation experiment does not extract a static knowledge graph for keyframes, so the results of knowledge graph are not inputted into the prompts of static hallucination and dynamic hallucination, and other parts remain unchanged. For \textbf{w/o dynamic KG}, the ablation experiment does not extract a dynamic knowledge graph for keyframe clusters and the keyframe group, so the results of the knowledge graph are not inputted into the prompt of dynamic hallucination while other parts remain unchanged. The results of the ablation experiments are presented in Table \ref{main_results}. These experiments provide insights into the effectiveness of the knowledge graph construction and its role in static and dynamic hallucination detection.

\section{Conclusion}

In conclusion, our work addresses the critical challenge of video hallucination detection in large T2V language models like Sora. By leveraging a combination of keyframe extraction, object detection, knowledge graph construction, and multimodal large language models, our methodology provides a comprehensive solution for identifying both static and dynamic hallucinations. The introduction of a benchmark dataset further supports the development and evaluation of hallucination detection methods, providing a valuable resource for the research community. Our experimental results demonstrate the superior performance of our approach, highlighting its effectiveness and potential for real-world applications.

As large T2V language models like Sora continue to advance and find applications in diverse fields, developing robust methods for detecting and mitigating hallucinations becomes increasingly critical to ensure the responsible and reliable use of these powerful models. Our work contributes to improving the quality and reliability of these models in video generation tasks, fostering trust in their applications and opening avenues for further research and advancements in this field.

\newpage

\bibliographystyle{unsrt}
\bibliography{sora_detector}

\appendix

\section{Appendix - Reports}



\section*{Supplementary material}

**Report 1**:

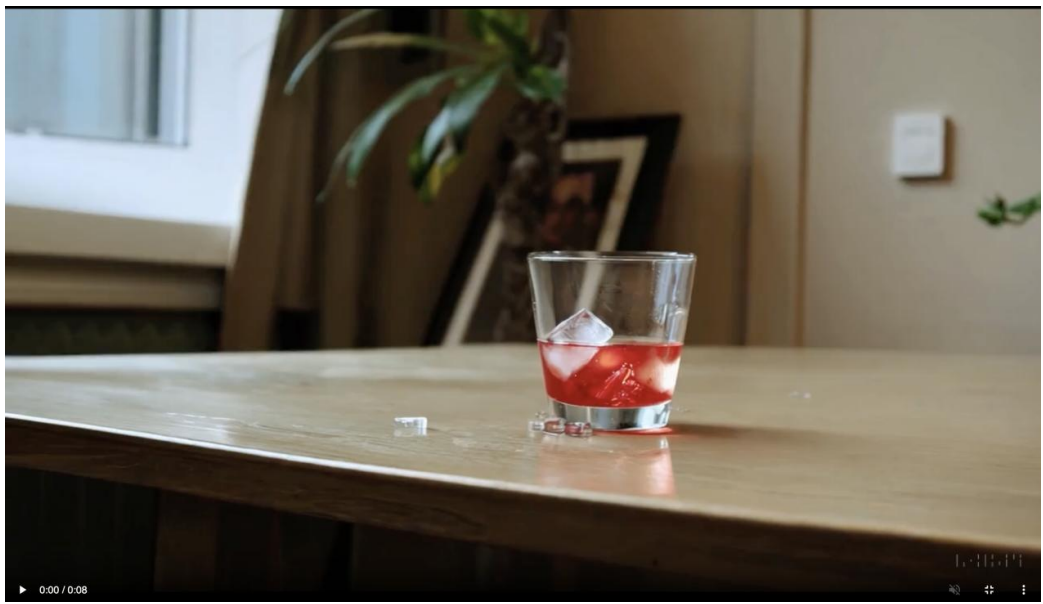

**Video Length:** 8s

**Video Summary:** The video illustrates the sequence of a glass tipping over on a wooden table. Initially, the glass is upright and contains ice cubes and red liquid. As the sequence progresses, the glass begins to tilt more drastically. Throughout, the ice cubes remain relatively stable within the glass, but the red liquid spills progressively, increasing in area on the table as the glass reaches a full tilt and eventually lies on its side. The table remains unchanged throughout the sequence.

**<Option>Original Prompt:** Pouring red liquid onto the table.

**Hallucination Detected:**

- Inconsistency with the original prompt.

- The edges of the ice cube inside the glass seem slightly too sharp and lacking slight melting. 3.68s

- The glass went from lying to standing without any depicted external force aiding this transition. 5.00s ~ 5.66s

- The red liquid is being dispensed from the bottom of the cup, causing an implausible penetration. 5.00s ~ 5.66s

- The glass falls down from air to the table without external force. 7.19s ~ 7.94s

**Overall Video Quality Score:** 55.6

**Detection Detail:**

1.key frame and detail frame extraction:

The number of key frame: 3

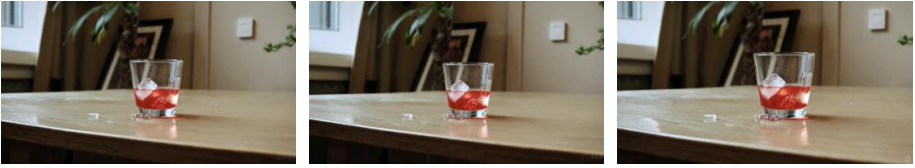

key_frame_1(center): 3.68s
detail_frame_l1(left): 1.33s
detail_frame_r1(right): 4.57s

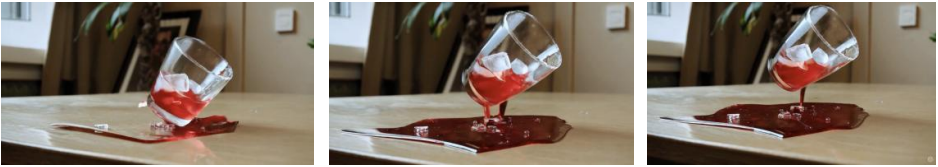

key_frame_2(center): 5.51s
detail_frame_l1(left): 5.00s
detail_frame_r1(right): 5.66s

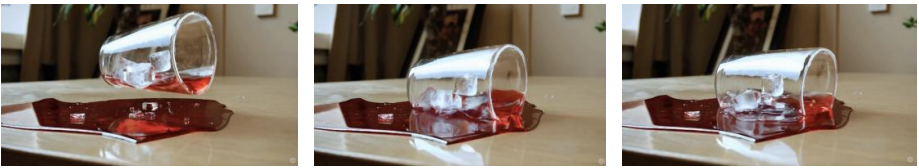

key_frame_3(center): 7.73s
detail_frame_l1(left): 7.19s
detail_frame_r1(right): 7.94s

2.object extraction and static KG construction for each key frame:

key_frame_1: {{"objects": ["glass", "ice_cube", "red_liquid"], "relations": [["ice_cube", "inside", "glass"], ["red_liquid", "inside", "glass"]]}}

key_frame_2:{"objects": ["glass", "ice_cube", "red_liquid", "table", "spilled_red_liquid"], "relations": [["glass", "contains", "ice_cube"], [["glass", "contains", "red_liquid"], ["glass", "pouring", "red_liquid"], ["spilled_red_liquid", "on", "table"]]}

key_frame_3:{{"objects": ["glass", "ice_cube", "red_liquid"], "relations": [["ice_cube", "inside", "glass"], ["red_liquid", "under", "glass"]]}}

3.object change extraction for each key frame cluster:

key_frame_cluster_1: {"changes":{"glass": "unchanged", "ice_cube": "unchanged", "red_liquid": "unchanged"}}

key_frame_cluster_2:{"changes":{"glass":"changed tilt progressively","ice_cube": "unchanged","red_liquid":"spilled progressively", "table ":"unchanged","spilled red liquid":"increased area progressively"}}

key_frame_cluster_3: {"changes":{"glass": "from the air to the table", "ice_cube": "unchanged", "red_liquid": "unchanged"}}

4.video summary:

Summary: The video illustrates the sequence of a glass tipping over on a wooden table. Initially, the glass is upright and contains ice cubes and red liquid. As the sequence progresses, the glass begins to tilt more drastically. Throughout, the ice cubes remain relatively stable within the glass, but the red liquid spills

progressively, increasing in area on the table as the glass reaches a full tilt and eventually lies on its side. The table remains unchanged throughout the sequence.

5.inconsistency comparison:

Analysis: Analyzing the provided key frames and comparing them against the prompts, the following discrepancies are noted: 1. **Glass Placement and Action**, 2. **Action Source**. These inconsistencies primarily revolve around the nature of the action (passive spill vs. active pour) and the agent of action (implicitly self-tipping glass vs. potentially human-caused pouring).

Inconsistency Score: 7

6.static hallucination detection:

**key_frame_1:**

| Hallucination Type | Hallucination Analysis | Hallucination Score |
| --- | --- | --- |
| Geometric Structure Irrationality | The edges of the ice cube inside the glass seem slightly too sharp and lacking slight melting. | 1 |
| Biological Structure Irrationality | / | 0 |
| Lighting, Shadow and Material Physical Inaccuracy | Slight texture anomaly potentially in the perspective or edges of the ice in contact with liquid. | 2 |
| Color Distribution Disharmony | / | 0 |
| Depth of Field and Focal Length Unreality | / | 0 |
| Object Composition and Scene Semantic Inconsistency | / | 0 |
| Motion and Blur Inconsistency | / | 0 |
| Physical Phenomenon Inauthenticity | / | 0 |
| Image Quality Inconsistency | / | 0 |
| Other Hallucination | / | 0 |

**key_frame_2:**

| Hallucination Type | Hallucination Analysis | Hallucination Score |
| --- | --- | --- |
| Geometric Structure Irrationality | / | 0 |
| Biological Structure Irrationality | / | 0 |
| Lighting, Shadow and Material Physical Inaccuracy | / | 0 |
| Color Distribution Disharmony | / | 0 |
| Depth of Field and Focal | / | 0 |

| Length Unreality | | |
| Object Composition and Scene Semantic Inconsistency | / | 0 |
| Motion and Blur Inconsistency | / | 0 |
| Physical Phenomenon Inauthenticity | / | 0 |
| Image Quality Inconsistency | / | 0 |
| Other Hallucination | / | 0 |

**key_frame_3:**

| Hallucination Type | Hallucination Analysis | Hallucination Score |
| --- | --- | --- |
| Geometric Structure Irrationality | / | 0 |
| Biological Structure Irrationality | / | 0 |
| Lighting, Shadow and Material Physical Inaccuracy | / | 0 |
| Color Distribution Disharmony | / | 0 |
| Depth of Field and Focal Length Unreality | / | 0 |
| Object Composition and Scene Semantic Inconsistency | / | 0 |
| Motion and Blur Inconsistency | / | 0 |
| Physical Phenomenon Inauthenticity | / | 0 |
| Image Quality Inconsistency | / | 0 |
| Other Hallucination | / | 0 |

7.dynamic hallucination detection:

**key_frame_cluster_1:**

| Hallucination Type | Hallucination Analysis | Hallucination Score |
| --- | --- | --- |
| Clipping | / | 0 |
| Implausible Fusion | / | 0 |
| Implausible Appearance or Disappearance | / | 0 |
| Implausible Motion | / | 0 |
| Implausible Transform | / | 0 |
| Implausible Penetration | / | 0 |
| Physical Interaction Errors | / | 0 |
| Logical Interaction Error | / | 0 |

| | | |
|---|---|---|
| Other Hallucination | / | 0 |

**key_frame_cluster_2:**

| Hallucination Type | Hallucination Analysis | Hallucination Score |
|---|---|---|
| Clipping | / | 0 |
| Implausible Fusion | / | 0 |
| Implausible Appearance or Disappearance | / | 0 |
| Implausible Motion | The glass went from lying to standing without any depicted external force aiding this transition. | 9 |
| Implausible Transform | / | 0 |
| Implausible Penetration | The red liquid is being dispensed from the bottom of the cup, causing an implausible penetration. | 8 |
| Physical Interaction Errors | / | 0 |
| Logical Interaction Error | / | 0 |
| Other Hallucination | / | 0 |

**key_frame_cluster_3:**

| Hallucination Type | Hallucination Analysis | Hallucination Score |
|---|---|---|
| Clipping | / | 0 |
| Implausible Fusion | / | 0 |
| Implausible Appearance or Disappearance | / | 0 |
| Implausible Motion | The glass falls down from air to the table without external force, shows the obvious phenomenon of Implausible Motion Hallucination. | 7 |
| Implausible Transform | / | 0 |
| Implausible Penetration | / | 0 |
| Physical Interaction Errors | / | 0 |
| Logical Interaction Error | / | 0 |
| Other Hallucination | / | 0 |

8.aggregation and overall scoring:

Overall Hallucination Score Aggregation:

| Inconsistency Score | Static Hallucination Score | Dynamic Hallucination Score |
|---|---|---|
| 7 | 1.73 | 5.87 |

Overall Score: 100 - 2 * 7 - 4 * 1.73 - 4 * 5.87 = 55.6

**Report 2**:

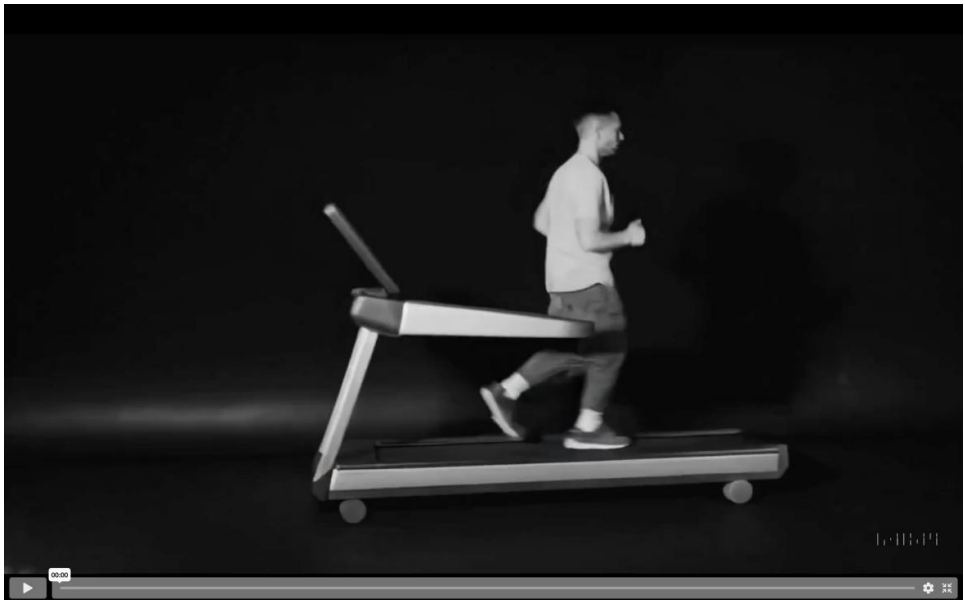

**Video Length:** 20s
**Video Summary:** The video shows a man continuously running on a treadmill in a monochrome setting. The treadmill remains unchanged throughout the video, indicating that the focus is on the man's steady exercise routine. The backdrop is dark, emphasizing the solitary figure as he maintains his pace on the treadmill.
**<Option>Original Prompt:** Step-printing scene of a person running, cinematic film shot in 35mm.
**Hallucination Detected:**
- Inconsistency with the original prompt.
- The person depicted appears to be running on a treadmill in what seems like the opposite direction to which one typically runs on such exercise equipment, giving an unusual and seemingly illogical appearance.
**Overall Video Quality Score:** 56

**Detection Detail:**

<span style="color:red">1.key frame and detail frame extraction:</span>

The number of key frame: 1

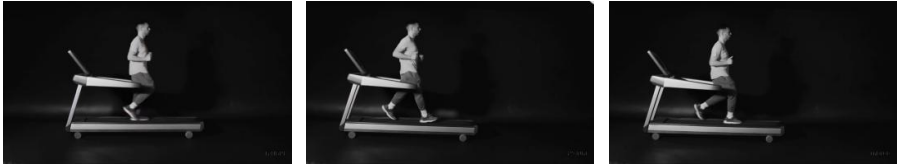

key_frame_1(center): 8.96s
detail_frame_l1(left): 7.71s
detail_frame_r1(right): 12.10s

<span style="color:red">2.object extraction and static KG construction for each key frame:</span>

key_frame_1:{{"objects": ["man", "treadmill"], "relations":[["man", "running on", "treadmill"]]}}

<span style="color:orange">3.object change extraction for each key frame cluster:</span>

key_frame_cluster_1: {"changes":{"man": "continuously maintains running", "treadmill": "unchanged"}}

<span style="color:green">4.video summary:</span>

Summary: The video shows a man continuously running on a treadmill in a monochrome setting. The treadmill remains unchanged throughout the video, indicating that the focus is on the man's steady exercise routine. The backdrop is dark, emphasizing the solitary figure as he maintains his pace on the treadmill.

<span style="color:green">5.inconsistency comparison:</span>

Analysis: The main inconsistency lies in the interpretation of the visual presentation and intention behind the video. While the original prompt implies a potential artistic or filmic exploration using step-printing (a process that could alter perception of motion and time), the generated prompt suggests a straightforward, unaltered depiction of a man running on a treadmill in a monochrome setting. The specifics about the monochrome effect and the clear focus on the exercise routine in the generated prompt contrast with the more technique-focused and possibly more dynamically edited implication of the origin prompt. This inconsistency, though significant in the context of film technique versus straightforward depiction, is moderated by the consistency in still describing a person running, hence the middling score.
Inconsistency Score: 5

<span style="color:cyan">6.static hallucination detection:</span>

**key_frame_1:**

| Hallucination Type | Hallucination Analysis | Hallucination Score |
|---|---|---|
| Geometric Structure Irrationality | / | 0 |
| Biological Structure Irrationality | / | 0 |
| Lighting, Shadow and Material Physical Inaccuracy | / | 0 |
| Color Distribution Disharmony | / | 0 |
| Depth of Field and Focal Length Unreality | / | 0 |
| Object Composition and | / | 0 |

| Scene Semantic Inconsistency | | |
| --- | --- | --- |
| Motion and Blur Inconsistency | / | 0 |
| Physical Phenomenon Inauthenticity | / | 0 |
| Image Quality Inconsistency | / | 0 |
| Other Hallucination | / | 0 |

## 7.dynamic hallucination detection:

**key_frame_cluster_1:**

| Hallucination Type | Hallucination Analysis | Hallucination Score |
| --- | --- | --- |
| Clipping | / | 0 |
| Implausible Fusion | / | 0 |
| Implausible Appearance or Disappearance | / | 0 |
| Implausible Motion | The person depicted appears to be running on a treadmill in what seems like the opposite direction to which one typically runs on such exercise equipment, giving an unusual and seemingly illogical appearance. | 8 |
| Implausible Transform | / | 0 |
| Implausible Penetration | / | 0 |
| Physical Interaction Errors | / | 0 |
| Logical Interaction Error | / | 0 |
| Other Hallucination | / | 0 |

## 8.aggregation and overall scoring:

Overall Hallucination Score Aggregation:

| Inconsistency Score | Static Hallucination Score | Dynamic Hallucination Score |
| --- | --- | --- |
| 5 | 0 | 8 |

Overall Score: 100 - 2 * 5 - 4 * 0 - 4 * 8 = 56

**Report 3:**

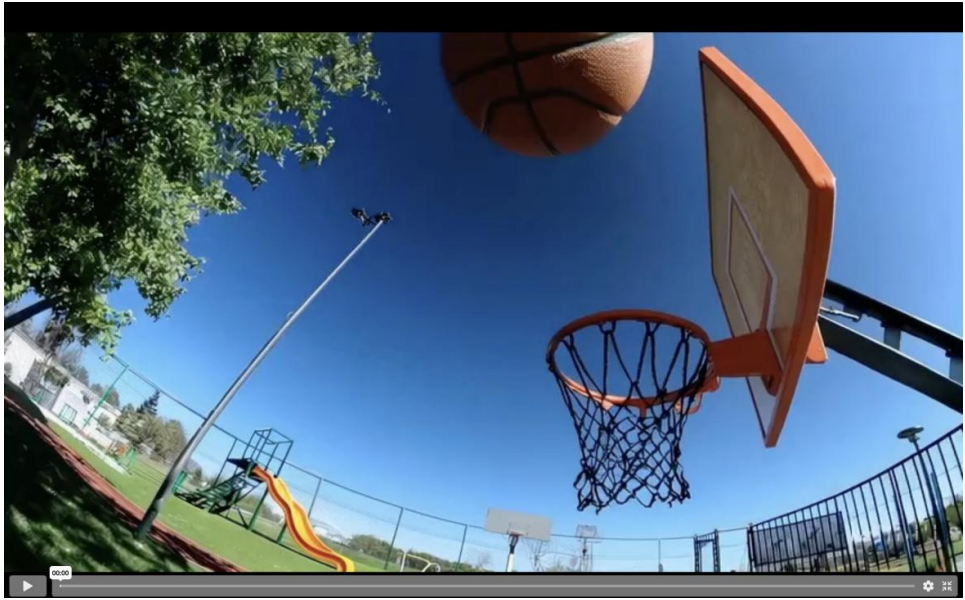

**Video Length:** 8s

**Video Summary:** The video showcases a sequence of a basketball moving towards and then through a hoop. Initially, the basketball is seen flying towards the hoop. As it descends, the ball suddenly ignites, adding a dramatic effect with flames visible in the backdrop, providing a visual enhancement rather than an interaction with the surroundings or the ball itself. The ball then continues its path downwards through the hoop, eventually passing through the net without altering its structure. Throughout these events, the surroundings such as the basketball hoop, backboard, net, light pole, playground slide, and fence remain static and unchanged. This dynamic display contrasts the stationary background with the fiery motion of the basketball.

**<Option>Original Prompt:** Basketball through hoop then explodes.

**Hallucination Detected:**

- Inconsistency with the original prompt.
- The basketball ignites into flames which is highly implausible and represents a severe hallucination. A basketball normally wouldn't ignite while playing. 4.80s ~ 5.60s
- The ignition of the basketball could imply a physical interaction error since such behavior (ignition without a plausible source) doesn't align with normal physical interactions. 4.80s ~ 5.60s
- There is a sudden appearance of basketball near the Basketball hoop between frame 2 and 3 without any visible cause. 6.08s ~ 6.72s
- The basketball meets an unnatural overlapping and intersection with basketball hoop, indicating the hallucination of clipping. 7.52s ~ 7.94s

**Overall Video Quality Score:** 55.16

**Detection Detail:**

1.key frame and detail frame extraction:

The number of key frame: 4

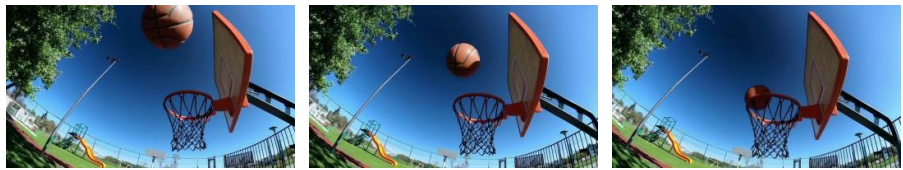

key_frame_1(center): 1.12s
detail_frame_l1(left): 0.48s
detail_frame_r1(right): 1.6s

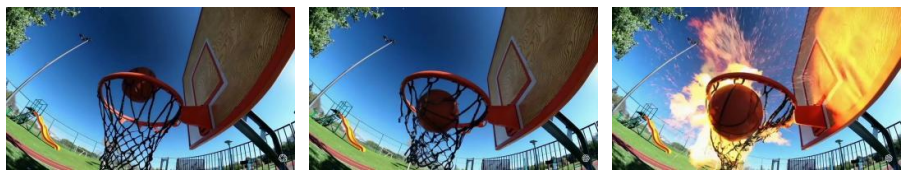

key_frame_2(center): 5.28s
detail_frame_l1(left): 4.80s
detail_frame_r1(right): 5.60s

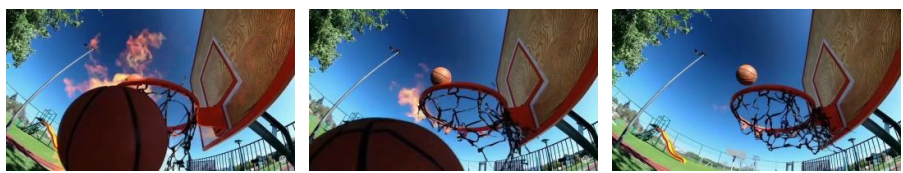

key_frame_3(center): 6.40s
detail_frame_l1(left): 6.08s
detail_frame_r1(right): 6.72s

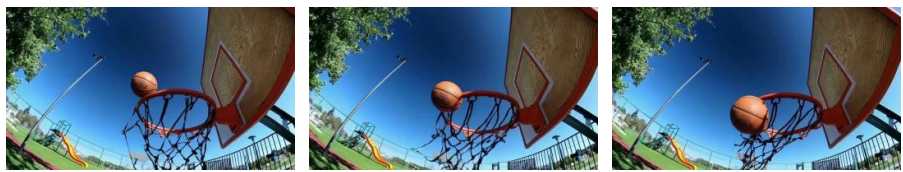

key_frame_4(center): 7.84s
detail_frame_l1(left): 7.52s
detail_frame_r1(right): 7.94s

2.object extraction and static KG construction for each key frame:

key_frame_1:{{"objects": ["basketball_1", "basketball_hoop", "backboard", "light_pole", "playground_slide"], "relations":[["basketball_1", "approaching", "basketball_hoop"], ["basketball_1", "below", "backboard"], ["light_pole", "adjacent to", "basketball_hoop"], ["playground_slide", "near", "basketball_hoop"]]}}

key_frame_2:{{"objects": ["basketball_hoop", "basketball", "backboard", "basketball_net", "light_pole", slide"], "relations":[["basketball", "inside", "basketball_net"], ["basketball_net", "attached_to", "backboard"], ["backboard", "attached_to", "basketball_hoop"]]}}

key_frame_3:{{"objects": ["Basketball_1", "Basketball_2", "Basketball hoop", "Flame"], "relations":[["Basketball_1", "approaching", "Basketball hoop"], ["Basketball_2", "inside", "Basketball hoop"], ["Flame", "behind", "Basketball hoop"]]}}

key_frame_4:{"objects": ["basketball", "basketball_hoop", "basketball_net", "light_pole", "slide", "fence"],"relations": [["basketball", "entering", "basketball_net"],["basketball_net", "attached to", "basketball_hoop"],["basketball_hoop", "mounted on", "backboard"],["slide", "located at",

"playground"],["fence", "surrounds", "basketball_court"]]}

3.object change extraction for each key frame cluster:
key_frame_cluster_1:{"changes":{"basketball_1": "moving towards the basketball hoop", "basketball_hoop": "unchanged", "backboard": "unchanged", "light_pole": "unchanged", "playground_slide": "unchanged"}}
key_frame_cluster_2:{"changes":{"basketball_hoop": "unchanged", "basketball": "moves downward, then ignites", "backboard": "unchanged", "basketball_net": "unchanged in structure but interacts with basketball", "light_pole": "unchanged", "slide": "unchanged"}}
key_frame_cluster_3:{"changes":{"Basketball_1": "consistently approaching Basketball hoop", "Basketball_2": "unchanged inside Basketball hoop", "Basketball hoop": "unchanged", "Flame": "consistent, natural variation behind Basketball hoop"}}
key_frame_cluster_4:{{"changes": {"basketball": "moves downward through the net","basketball_hoop": "unchanged", "basketball_net": "unchanged", "light_pole": "unchanged", "slide": "unchanged", "fence": "unchanged"}}}

4.video summary:
Summary: The video showcases a sequence of a basketball moving towards and then through a hoop. Initially, the basketball is seen flying towards the hoop. As it descends, the ball suddenly ignites, adding a dramatic effect with flames visible in the backdrop, providing a visual enhancement rather than an interaction with the surroundings or the ball itself. The ball then continues its path downwards through the hoop,eventually passing through the net without altering its structure. Throughout these events, the surroundings such as the basketball hoop, backboard, net, light pole, playground slide, and fence remain static and unchanged. This dynamic display contrasts the stationary background with the fiery motion of the basketball.

5.inconsistency comparison:
Analysis: The discrepancy lies in the interpretation of "explodes." The origin prompt suggests a literal explosion occurs post-hoop interaction, whereas the generated video and its description portray the basketball catching on fire without disintegrating, which is less severe than an explosion.
Inconsistency Score: 6

6.static hallucination detection:
**key_frame_1:**

| Hallucination Type | Hallucination Analysis | Hallucination Score |
|---|---|---|
| Geometric Structure Irrationality | / | 0 |
| Biological Structure Irrationality | / | 0 |
| Lighting, Shadow and Material Physical Inaccuracy | / | 0 |
| Color Distribution Disharmony | / | 0 |
| Depth of Field and Focal Length Unreality | / | 0 |
| Object Composition and Scene Semantic Inconsistency | / | 0 |
| Motion and Blur Inconsistency | / | 0 |

| Physical Phenomenon Inauthenticity | / | 0 |
| Image Quality Inconsistency | / | 0 |
| Other Hallucination | / | 0 |

**key_frame_2:**

| Hallucination Type | Hallucination Analysis | Hallucination Score |
| --- | --- | --- |
| Geometric Structure Irrationality | / | 0 |
| Biological Structure Irrationality | / | 0 |
| Lighting, Shadow and Material Physical Inaccuracy | / | 0 |
| Color Distribution Disharmony | / | 0 |
| Depth of Field and Focal Length Unreality | / | 0 |
| Object Composition and Scene Semantic Inconsistency | / | 0 |
| Motion and Blur Inconsistency | / | 0 |
| Physical Phenomenon Inauthenticity | / | 0 |
| Image Quality Inconsistency | / | 0 |
| Other Hallucination | / | 0 |

**key_frame_3:**

| Hallucination Type | Hallucination Analysis | Hallucination Score |
| --- | --- | --- |
| Geometric Structure Irrationality | / | 0 |
| Biological Structure Irrationality | / | 0 |
| Lighting, Shadow and Material Physical Inaccuracy | / | 0 |
| Color Distribution Disharmony | / | 0 |
| Depth of Field and Focal Length Unreality | / | 0 |
| Object Composition and Scene Semantic Inconsistency | / | 0 |
| Motion and Blur Inconsistency | / | 0 |

| Physical Phenomenon Inauthenticity | / | 0 |
| Image Quality Inconsistency | / | 0 |
| Other Hallucination | / | 0 |

**key_frame_4:**

| Hallucination Type | Hallucination Analysis | Hallucination Score |
|---|---|---|
| Geometric Structure Irrationality | / | 0 |
| Biological Structure Irrationality | / | 0 |
| Lighting, Shadow and Material Physical Inaccuracy | / | 0 |
| Color Distribution Disharmony | / | 0 |
| Depth of Field and Focal Length Unreality | / | 0 |
| Object Composition and Scene Semantic Inconsistency | / | 0 |
| Motion and Blur Inconsistency | / | 0 |
| Physical Phenomenon Inauthenticity | / | 0 |
| Image Quality Inconsistency | / | 0 |
| Other Hallucination | / | 0 |

7.dynamic hallucination detection:

**key_frame_cluster_1:**

| Hallucination Type | Hallucination Analysis | Hallucination Score |
|---|---|---|
| Clipping | / | 0 |
| Implausible Fusion | / | 0 |
| Implausible Appearance or Disappearance | / | 0 |
| Implausible Motion | / | 0 |
| Implausible Transform | / | 0 |
| Implausible Penetration | / | 0 |
| Physical Interaction Errors | / | 0 |
| Logical Interaction Error | / | 0 |
| Other Hallucination | / | 0 |

**key_frame_cluster_2:**

| Hallucination Type | Hallucination Analysis | Hallucination Score |
|---|---|---|
| Clipping | / | 0 |
| Implausible Fusion | / | 0 |

| Implausible Appearance or Disappearance | / | 0 |
|---|---|---|
| Implausible Motion | / | 0 |
| Implausible Transform | The basketball ignites into flames which is highly implausible and represents a severe hallucination. A basketball normally wouldn't ignite while playing. | 10 |
| Implausible Penetration | / | 0 |
| Physical Interaction Errors | The ignition of the basketball could imply a physical interaction error since such behavior (ignition without a plausible source) doesn't align with normal physical interactions. | 8 |
| Logical Interaction Error | / | 0 |
| Other Hallucination | / | 0 |

**key_frame_cluster_3:**

| Hallucination Type | Hallucination Analysis | Hallucination Score |
|---|---|---|
| Clipping | / | 0 |
| Implausible Fusion | / | 0 |
| Implausible Appearance or Disappearance | There is a sudden appearance of basketball near the Basketball hoop between frame 2 and 3 without any visible cause. | 8 |
| Implausible Motion | / | 0 |
| Implausible Transform | / | 0 |
| Implausible Penetration | / | 0 |
| Physical Interaction Errors | / | 0 |
| Logical Interaction Error | / | 0 |
| Other Hallucination | / | 0 |

**key_frame_cluster_4:**

| Hallucination Type | Hallucination Analysis | Hallucination Score |
|---|---|---|
| Clipping | The basketball meets an unnatural overlapping and intersection with basketball hoop, indicating the hallucination of clipping. | 9 |
| Implausible Fusion | / | 0 |
| Implausible Appearance or Disappearance | / | 0 |
| Implausible Motion | / | 0 |
| Implausible Transform | / | 0 |
| Implausible Penetration | / | 0 |
| Physical Interaction Errors | / | 0 |
| Logical Interaction Error | / | 0 |
| Other Hallucination | / | 0 |

8.aggregation and overall scoring:

Overall Hallucination Score Aggregation:

| Inconsistency Score | Static Hallucination Score | Dynamic Hallucination Score |
|---|---|---|
| 6 | 0 | 8.21 |

Overall Score: 100 - 2 * 6 - 4 * 0 - 4 * 8.21 = 55.16

**Report 4**:

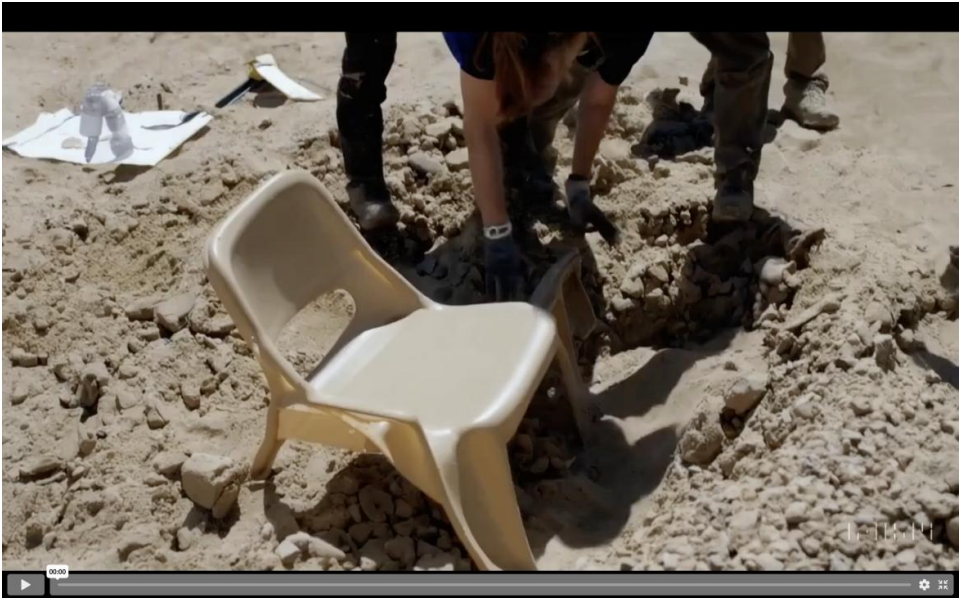

**Video Length:** 20s

**Video Summary:** The sequence of images depicts a group of individuals, possibly archaeologists or researchers, engaged in the process of unearthing an object buried in sandy terrain. The process is methodical and involves multiple stages:1. The initial frame shows two people starting to clear sand around the buried object using tools.2. Subsequent frames illustrate the gradual exposure of the object, which appears to be a plastic chair, as more sand is removed by the participants.3. One frame captures a moment where the chair is being lifted from the hole, indicating that it was successfully uncovered.4. Towards the conclusion of the sequence, the chair is fully removed from the ground, and the team inspects it, suggesting a sense of completion or discovery.Overall, the video likely showcases a controlled demonstration or experiment, possibly related to archaeology or geological studies, using everyday objects as practice or illustrative materials.

**<Option>Original Prompt:** Archeologists discover a generic plastic chair in the desert, excavating and dusting it with great care.

**Hallucination Detected:**

- Inconsistency with the original prompt.

- The shape presented by the chair resembles a fluid, which contradicts the fact that the chair is a hard object in reality. 3.67s ~ 8.33s

- The chair moves without external force, appearing to float in the air. 12.33s ~ 19.83s

**Overall Video Quality Score:** 58.52

**Detection Detail:**

1.key frame and detail frame extraction:

The number of key frame: 5

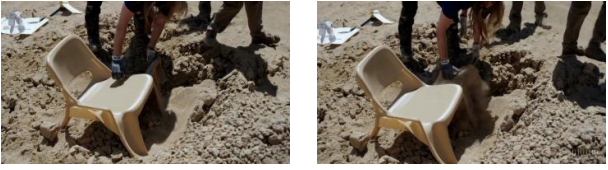

key_frame_1(left): 1.67s
detail_frame_r1(right): 2.33s

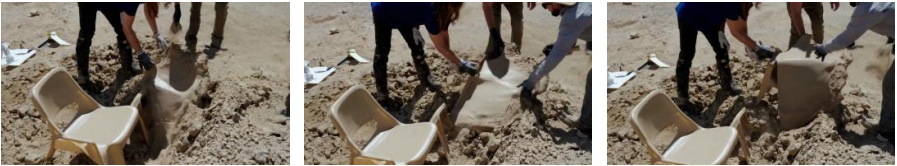

key_frame_2(center): 5.33s
detail_frame_l1(left): 3.67s
detail_frame_r1(right): 5.83s

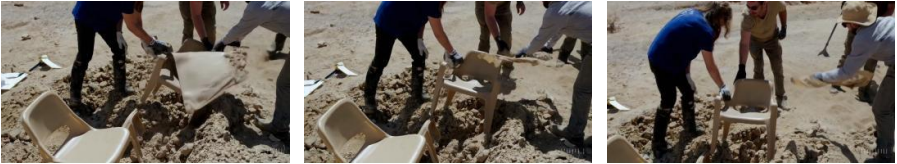

key_frame_3(center): 6.83s
detail_frame_l1(left): 6.17s
detail_frame_r1(right): 8.33s

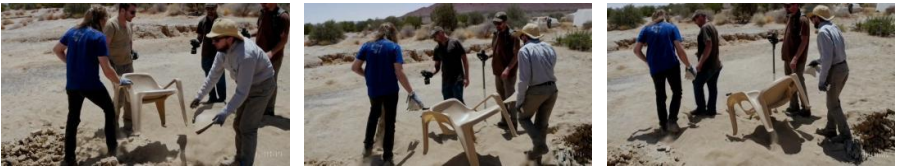

key_frame_4(center): 14.5s
detail_frame_l1(left): 12.33s
detail_frame_r1(right): 15.83s

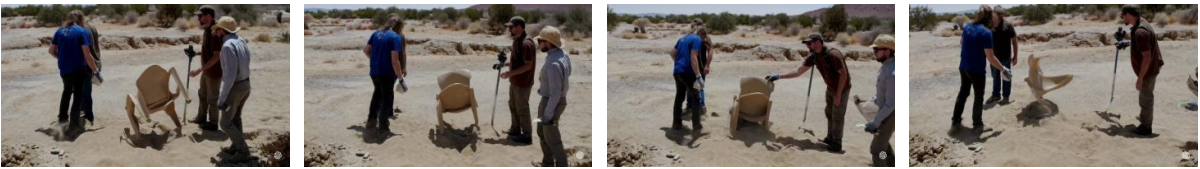

key_frame_5(third): 18.5s
detail_frame_l2(left): 16.83s
detail_frame_l1(second): 17.67s
detail_frame_r1(right): 19.83s

2.object extraction and static KG construction for each key frame:

key_frame_1:{{"objects": ["chair", "shovel_1", "shovel_2", "rock_1", "rock_2", "rock_3"], "relations":[["shovel_1", "used by", "person_1"], ["shovel_2", "used by", "person_2"], ["rock_1", "near", "chair"], ["rock_2", "near", "chair"], ["rock_3", "near", "chair"]]}}

key_frame_2:{{"objects": ["Plastic_Chair_1", "Plastic_Chair_2", "Shovel_1", "Shovel_2", "Plastic_Bucket"],

"relations":[["Shovel_1", "near", "Plastic_Chair_1"], ["Shovel_2", "is being used by", "Person_1"], ["Person_1", "is digging with", "Shovel_2"], ["Plastic_Chair_2", "near", "Person_2"], ["Person_2", "is standing near", "Plastic_Chair_2"], ["Plastic_Bucket", "is in front of", "Plastic_Chair_1"]]}}

key_frame_3:{{"objects": ["Chair_1", "Chair_2", "Dirt Pile"], "relations": [["Chair_1", "next to", "Chair_2"], ["Chair_1", "on", "Dirt Pile"], ["Chair_2", "on", "Dirt Pile"]]}}

key_frame_4:{{"objects": ["plastic_chair", "man_1", "man_2", "man_3", "camera_1", "camera_2"], "relations":[["man_1", "holding", "plastic_chair"], ["man_1", "walking towards", "man_2"], ["man_2", "interacting with", "man_3"], ["man_2", "holding", "camera_1"], ["man_3", "carrying", "camera_2"]]}}

key_frame_5:{{"objects": ["Person_1", "Person_2", "Person_3", "Person_4", "Armchair", "Tool"], "relations":[["Person_2", "reaching towards", "Armchair"], ["Person_3", "holding", "Tool"]]}}

### 3.object change extraction for each key frame cluster:

key_frame_cluster_1:{"changes":{"chair": "unchanged", "shovel_1": "used in continuous digging action", "shovel_2": "used in continuous digging action", "rock_1": "unchanged", "rock_2": "unchanged", "rock_3": "unchanged"}}

key_frame_cluster_2:{"changes": {"Plastic_Chair_1": "minor repositioning", "Plastic_Chair_2": "slight movement", "Shovel_1": "no change/unchanged", "Shovel_2": "motion consistent with digging", "Plastic_Bucket": "unchanged"}}

key_frame_cluster_3:{"changes":{"Chair_1": "moved from down to up", "Chair_2": "unchanged", "Dirt Pile": "unchanged"}}

key_frame_cluster_4:{"changes":{"plastic_chair": "changed position and orientation", "man_1": "changed position minimally, constant interaction with plastic_chair", "man_2": "minimal movement and orientation change, constant with camera_1", "man_3": "minimal orientation adjustments, constant with camera_2", "camera_1": "unchanged", "camera_2": "unchanged"}}

key_frame_cluster_5:{"changes":{"Person_1":"unchanged", "Person_2":"unchanged", "Person_3":"unchanged", "Person_4":"unchanged", "Armchair":"falls into the sand and flips upside down", "Tool":"unchanged"}}

### 4.video summary:

Summary: The sequence of images depicts a group of individuals, possibly archaeologists or researchers, engaged in the process of unearthing an object buried in sandy terrain. The process is methodical and involves multiple stages:1. The initial frame shows two people starting to clear sand around the buried object using tools.2. Subsequent frames illustrate the gradual exposure of the object, which appears to be a plastic chair, as more sand is removed by the participants.3. One frame captures a moment where the chair is being lifted from the hole, indicating that it was successfully uncovered.4. Towards the conclusion of the sequence, the chair is fully removed from the ground, and the team inspects it, suggesting a sense of completion or discovery.Overall, the video likely showcases a controlled demonstration or experiment, possibly related to archaeology or geological studies, using everyday objects as practice or illustrative materials.

### 5.inconsistency comparison:

Analysis: The core activities are similar, but the presentation and implied intent differ significantly, affecting the coherence when comparing the narratives laid out in both prompts.
Inconsistency Score: 6

### 6.static hallucination detection:

**key_frame_1:**

| Hallucination Type | Hallucination Analysis | Hallucination Score |
| --- | --- | --- |
| Geometric Structure Irrationality | / | 0 |
| Biological Structure | / | 0 |

| | | |
|---|---|---|
| Irrationality | | |
| Lighting, Shadow and Material Physical Inaccuracy | / | 0 |
| Color Distribution Disharmony | / | 0 |
| Depth of Field and Focal Length Unreality | / | 0 |
| Object Composition and Scene Semantic Inconsistency | / | 0 |
| Motion and Blur Inconsistency | / | 0 |
| Physical Phenomenon Inauthenticity | / | 0 |
| Image Quality Inconsistency | / | 0 |
| Other Hallucination | / | 0 |

**key_frame_2:**

| Hallucination Type | Hallucination Analysis | Hallucination Score |
|---|---|---|
| Geometric Structure Irrationality | / | 0 |
| Biological Structure Irrationality | / | 0 |
| Lighting, Shadow and Material Physical Inaccuracy | The shape presented by the chair resembles a fluid, which contradicts the fact that the chair is a hard object in reality. | 8 |
| Color Distribution Disharmony | / | 0 |
| Depth of Field and Focal Length Unreality | / | 0 |
| Object Composition and Scene Semantic Inconsistency | / | 0 |
| Motion and Blur Inconsistency | / | 0 |
| Physical Phenomenon Inauthenticity | / | 0 |
| Image Quality Inconsistency | / | 0 |
| Other Hallucination | / | 0 |

**key_frame_3:**

| Hallucination Type | Hallucination Analysis | Hallucination Score |
|---|---|---|
| Geometric Structure Irrationality | / | 0 |
| Biological Structure | / | 0 |

| Hallucination Type | Hallucination Analysis | Hallucination Score |
|---|---|---|
| Irrationality | | |
| Lighting, Shadow and Material Physical Inaccuracy | The shape presented by the chair resembles a fluid, which contradicts the fact that the chair is a hard object in reality. | 8 |
| Color Distribution Disharmony | / | 0 |
| Depth of Field and Focal Length Unreality | / | 0 |
| Object Composition and Scene Semantic Inconsistency | / | 0 |
| Motion and Blur Inconsistency | / | 0 |
| Physical Phenomenon Inauthenticity | / | 0 |
| Image Quality Inconsistency | / | 0 |
| Other Hallucination | / | 0 |

**key_frame_4:**

| Hallucination Type | Hallucination Analysis | Hallucination Score |
|---|---|---|
| Geometric Structure Irrationality | / | 0 |
| Biological Structure Irrationality | / | 0 |
| Lighting, Shadow and Material Physical Inaccuracy | / | 0 |
| Color Distribution Disharmony | / | 0 |
| Depth of Field and Focal Length Unreality | / | 0 |
| Object Composition and Scene Semantic Inconsistency | / | 0 |
| Motion and Blur Inconsistency | / | 0 |
| Physical Phenomenon Inauthenticity | / | 0 |
| Image Quality Inconsistency | / | 0 |
| Other Hallucination | / | 0 |

**key_frame_5:**

| Hallucination Type | Hallucination Analysis | Hallucination Score |
|---|---|---|
| Geometric Structure Irrationality | / | 0 |
| Biological Structure | / | 0 |

| | | |
|---|---|---|
| Irrationality | | |
| Lighting, Shadow and Material Physical Inaccuracy | / | 0 |
| Color Distribution Disharmony | / | 0 |
| Depth of Field and Focal Length Unreality | / | 0 |
| Object Composition and Scene Semantic Inconsistency | / | 0 |
| Motion and Blur Inconsistency | / | 0 |
| Physical Phenomenon Inauthenticity | / | 0 |
| Image Quality Inconsistency | / | 0 |
| Other Hallucination | / | 0 |

7.dynamic hallucination detection:

**key_frame_cluster_1:**

| Hallucination Type | Hallucination Analysis | Hallucination Score |
|---|---|---|
| Clipping | / | 0 |
| Implausible Fusion | / | 0 |
| Implausible Appearance or Disappearance | / | 0 |
| Implausible Motion | / | 0 |
| Implausible Transform | / | 0 |
| Implausible Penetration | / | 0 |
| Physical Interaction Errors | / | 0 |
| Logical Interaction Error | / | 0 |
| Other Hallucination | / | 0 |

**key_frame_cluster_2:**

| Hallucination Type | Hallucination Analysis | Hallucination Score |
|---|---|---|
| Clipping | / | 0 |
| Implausible Fusion | / | 0 |
| Implausible Appearance or Disappearance | / | 0 |
| Implausible Motion | / | 0 |
| Implausible Transform | / | 0 |
| Implausible Penetration | / | 0 |
| Physical Interaction Errors | / | 0 |
| Logical Interaction Error | / | 0 |
| Other Hallucination | / | 0 |

**key_frame_cluster_3:**

| Hallucination Type | Hallucination Analysis | Hallucination Score |
|---|---|---|

| Clipping | / | 0 |
| Implausible Fusion | / | 0 |
| Implausible Appearance or Disappearance | / | 0 |
| Implausible Motion | / | 0 |
| Implausible Transform | / | 0 |
| Implausible Penetration | / | 0 |
| Physical Interaction Errors | / | 0 |
| Logical Interaction Error | / | 0 |
| Other Hallucination | / | 0 |

**key_frame_cluster_4:**

| Hallucination Type | Hallucination Analysis | Hallucination Score |
| --- | --- | --- |
| Clipping | / | 9 |
| Implausible Fusion | / | 0 |
| Implausible Appearance or Disappearance | / | 0 |
| Implausible Motion | The chair moves without external force, appearing to float in the air. | 10 |
| Implausible Transform | / | 0 |
| Implausible Penetration | / | 0 |
| Physical Interaction Errors | / | 0 |
| Logical Interaction Error | / | 0 |
| Other Hallucination | / | 0 |

**key_frame_cluster_5:**

| Hallucination Type | Hallucination Analysis | Hallucination Score |
| --- | --- | --- |
| Clipping | / | 9 |
| Implausible Fusion | / | 0 |
| Implausible Appearance or Disappearance | / | 0 |
| Implausible Motion | The chair moves without external force, appearing to float in the air. | 9 |
| Implausible Transform | / | 0 |
| Implausible Penetration | / | 0 |
| Physical Interaction Errors | / | 0 |
| Logical Interaction Error | / | 0 |
| Other Hallucination | / | 0 |

8.aggregation and overall scoring:

Overall Hallucination Score Aggregation:

| Inconsistency Score | Static Hallucination Score | Dynamic Hallucination Score |
| --- | --- | --- |
| 6 | 2.87 | 4.5 |

Overall Score: 100 - 2 * 6 - 4 * 2.87 - 4 * 4.5 = 58.52

**Report 5**:

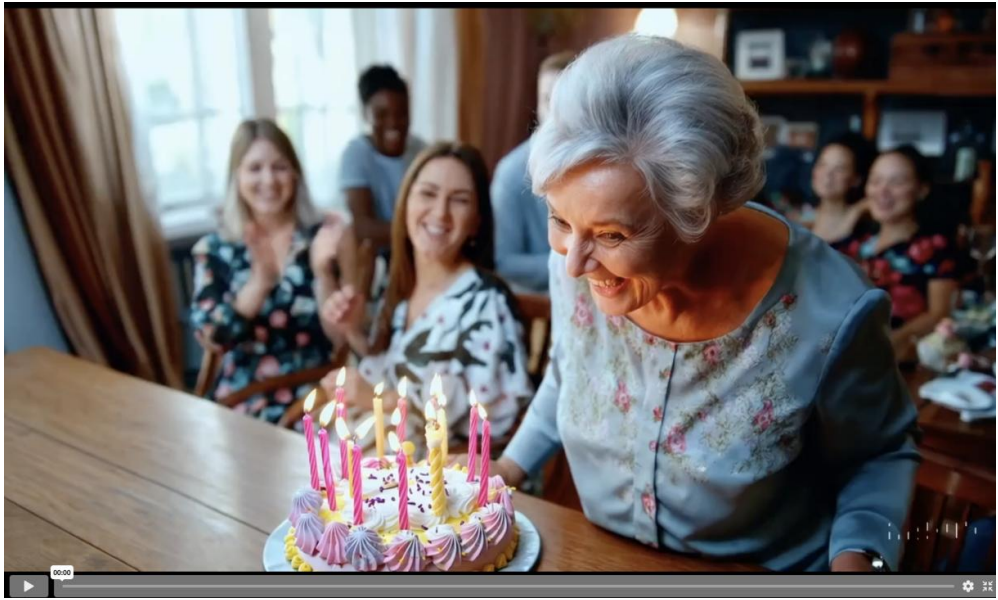

**Video Length:** 10s

**Video Summary:** The video captures a joyful celebration where an elderly woman is surrounded by a group of people at a birthday party. In the first key frame, she is presented with a colorful birthday cake adorned with lit candles, appearing delighted as she prepares to blow them out. The attendees in the background applaud and cheer for her. In the second key frame, she blows out the candles on her cake, marking the climax of the celebration as the guests continue to express their excitement and happiness. The atmosphere is festive, filled with laughter and applause, highlighting a warm, celebratory moment among friends and family.

**<Option>Original Prompt:** A grandmother with neatly combed grey hair stands behind a colorful birthday cake with numerous candles at a wood dining room table, expression is one of pure joy and happiness, with a happy glow in her eye. She leans forward and blows out the candles with a gentle puff, the cake has pink frosting and sprinkles and the candles cease to flicker, the grandmother wears a light blue blouse adorned with floral patterns, several happy friends and family sitting at the table can be seen celebrating, out of focus. The scene is beautifully captured, cinematic, showing a 3/4 view of the grandmother and the dining room. Warm color tones and soft lighting enhance the mood.

**Hallucination Detected:**

- Inconsistency with the original prompt.

- The woman is blowing out the candle but it has not been extinguished, presenting a physical phenomenon that should have occurred but has not occurred. 6.77s ~ 8.00s

**Overall Video Quality Score:** 72.68

**Detection Detail:**

1.key frame and detail frame extraction:

The number of key frame: 2

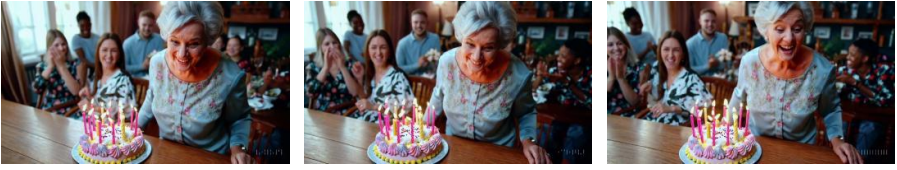

key_frame_1(center): 3.2s
detail_frame_l1(left): 1.53s
detail_frame_r1(right): 4.92s

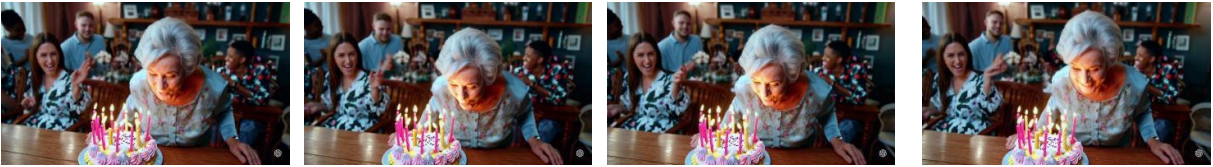

key_frame_2(second): 7.23s
detail_frame_l1(left): 6.77s
detail_frame_r1(third): 7.54s
detail_frame_r2(right): 8.00s

2.object extraction and static KG construction for each key frame:

key_frame_1:{{"objects": ["cake", "candle_1", "candle_2", "candle_3", "candle_4", "candle_5", "candle_6", "candle_7", "candle_8", "candle_9", "candle_10", "candle_11", "candle_12", "candle_13", "candle_14", "candle_15", "candle_16", "candle_17", "candle_18", "candle_19", "candle_20", "candle_21", "candle_22"], "relations":[["candle_1", "on", "cake"], ["candle_2", "on", "cake"], ["candle_3", "on", "cake"], ["candle_4", "on", "cake"], ["candle_5", "on", "cake"], ["candle_6", "on", "cake"], ["candle_7", "on", "cake"], ["candle_8", "on", "cake"], ["candle_9", "on", "cake"], ["candle_10", "on", "cake"], ["candle_11", "on", "cake"], ["candle_12", "on", "cake"], ["candle_13", "on", "cake"], ["candle_14", "on", "cake"], ["candle_15", "on", "cake"], ["candle_16", "on", "cake"], ["candle_17", "on", "cake"], ["candle_18", "on", "cake"], ["candle_19", "on", "cake"], ["candle_20", "on", "cake"], ["candle_21", "on", "cake"], ["candle_22", "on", "cake"]]}}

key_frame_2:{{"objects": ["woman", "birthday_cake", "candle_1", "candle_2", "candle_3", "candle_4", "candle_5", "candle_6", "candle_7", "candle_8", "candle_9", "candle_10", "candle_11", "people_group"], "relations":[["woman", "blowing", "candle_1"], ["woman", "blowing", "candle_2"], ["woman", "blowing", "candle_3"], ["woman", "blowing", "candle_4"], ["woman", "blowing", "candle_5"], ["woman", "blowing", "candle_6"], ["woman", "blowing", "candle_7"], ["woman", "blowing", "candle_8"], ["woman", "blowing", "candle_9"], ["woman", "blowing", "candle_10"], ["woman", "blowing", "candle_11"], ["birthday_cake", "adorned_with", "candle_1"], ["birthday_cake", "adorned_with", "candle_2"], ["birthday_cake", "adorned_with", "candle_3"], ["birthday_cake", "adorned_with", "candle_4"], ["birthday_cake", "adorned_with", "candle_5"], ["birthday_cake", "adorned_with", "candle_6"], ["birthday_cake", "adorned_with", "candle_7"], ["birthday_cake", "adorned_with", "candle_8"], ["birthday_cake", "adorned_with", "candle_9"], ["birthday_cake", "adorned_with", "candle_10"], ["birthday_cake", "adorned_with", "candle_11"]]}}

3.object change extraction for each key frame cluster:

key_frame_cluster_1:{"changes":{"cake": "unchanged", "candle_1": "unchanged", "candle_2": "unchanged", "candle_3": "unchanged", "candle_4": "unchanged", "candle_5": "unchanged", "candle_6": "unchanged", "candle_7": "unchanged", "candle_8": "unchanged", "candle_9": "unchanged", "candle_10": "unchanged", "candle_11": "unchanged", "candle_12": "unchanged", "candle_13": "unchanged", "candle_14": "unchanged", "candle_15": "unchanged", "candle_16": "unchanged", "candle_17": "unchanged", "candle_18": "unchanged",

"candle_19": "unchanged", "candle_20": "unchanged", "candle_21": "unchanged", "candle_22": "unchanged"}}
key_frame_cluster_2:{"changes":{"woman":"unchanged","birthday_cake":"unchanged","candle_1":"unchanged","candle_2":"unchanged","candle_3":"unchanged","candle_4":"unchanged","candle_5":"unchanged","candle_6":"unchanged","candle_7":"unchanged","candle_8":"unchanged","candle_9":"unchanged","candle_10":"unchanged","candle_11":"unchanged","people_group":"unchanged"}}

4.video summary:

Summary: The video captures a joyful celebration where an elderly woman is surrounded by a group of people at a birthday party. In the first key frame, she is presented with a colorful birthday cake adorned with lit candles, appearing delighted as she prepares to blow them out. The attendees in the background applaud and cheer for her. In the second key frame, she blows out the candles on her cake, marking the climax of the celebration as the guests continue to express their excitement and happiness. The atmosphere is festive, filled with laughter and applause, highlighting a warm, celebratory moment among friends and family.

5.inconsistency comparison:

Analysis: While there is a considerable overlap in the general theme and sequence of events, the origin prompt provides much more precise visual and emotional detail compared to the more generically described generated prompt. These missing details in the generated prompt lead to a noticeable inconsistency in depth and vividness of the scene portrayal.
Inconsistency Score: 6

6.static hallucination detection:

**key_frame_1:**

| Hallucination Type | Hallucination Analysis | Hallucination Score |
| --- | --- | --- |
| Geometric Structure Irrationality | / | 0 |
| Biological Structure Irrationality | / | 0 |
| Lighting, Shadow and Material Physical Inaccuracy | / | 0 |
| Color Distribution Disharmony | / | 0 |
| Depth of Field and Focal Length Unreality | / | 0 |
| Object Composition and Scene Semantic Inconsistency | / | 0 |
| Motion and Blur Inconsistency | / | 0 |
| Physical Phenomenon Inauthenticity | / | 0 |
| Image Quality Inconsistency | / | 0 |
| Other Hallucination | / | 0 |

**key_frame_2:**

| Hallucination Type | Hallucination Analysis | Hallucination Score |
| --- | --- | --- |

| Hallucination Type | Hallucination Analysis | Hallucination Score |
|---|---|---|
| Geometric Structure Irrationality | / | 0 |
| Biological Structure Irrationality | / | 0 |
| Lighting, Shadow and Material Physical Inaccuracy | / | 0 |
| Color Distribution Disharmony | / | 0 |
| Depth of Field and Focal Length Unreality | / | 0 |
| Object Composition and Scene Semantic Inconsistency | / | 0 |
| Motion and Blur Inconsistency | / | 0 |
| Physical Phenomenon Inauthenticity | / | 0 |
| Image Quality Inconsistency | / | 0 |
| Other Hallucination | / | 0 |

7.dynamic hallucination detection:

**key_frame_cluster_1:**

| Hallucination Type | Hallucination Analysis | Hallucination Score |
|---|---|---|
| Clipping | / | 0 |
| Implausible Fusion | / | 0 |
| Implausible Appearance or Disappearance | / | 0 |
| Implausible Motion | / | 0 |
| Implausible Transform | / | 0 |
| Implausible Penetration | / | 0 |
| Physical Interaction Errors | / | 0 |
| Logical Interaction Error | / | 0 |
| Other Hallucination | / | 0 |

**key_frame_cluster_2:**

| Hallucination Type | Hallucination Analysis | Hallucination Score |
|---|---|---|
| Clipping | / | 0 |
| Implausible Fusion | / | 0 |
| Implausible Appearance or Disappearance | / | 0 |
| Implausible Motion | / | 0 |
| Implausible Transform | | 0 |
| Implausible Penetration | / | 0 |
| Physical Interaction Errors | The woman is blowing out the candle but it has not been extinguished, presenting a physical phenomenon that should have occurred but has not occurred. | 8 |

| Logical Interaction Error | / | 0 |
| Other Hallucination | / | 0 |

8.aggregation and overall scoring:

Overall Hallucination Score Aggregation:

| Inconsistency Score | Static Hallucination Score | Dynamic Hallucination Score |
|---|---|---|
| 6 | 0 | 3.83 |

Overall Score: 100 - 2 * 6 - 4 * 0 - 4 * 3.83 = 72.68

**Report 6**:

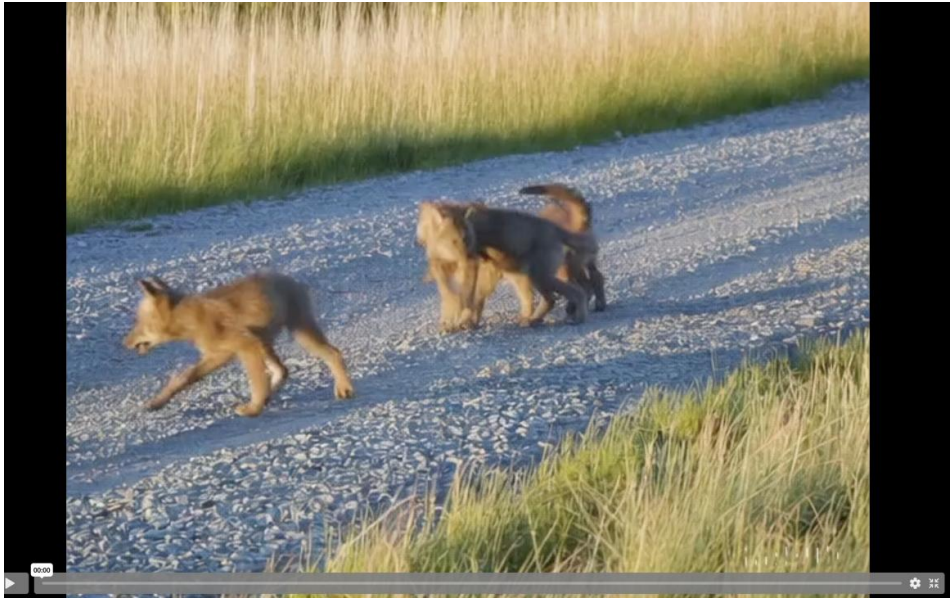

**Video Length:** 10s

**Video Summary:** The video depicts a playful scene featuring three young foxes on a gravel pathway surrounded by grassy areas. The sequence begins with one of the foxes darting across the path, characterized by its motion blur, indicating rapid movement. The next key frame shows the three foxes standing closely together in the middle of the path, playfully interacting with each other, as evidenced by their alert stances and physical proximity. In the final key frame, one fox remains active and engaged, with its tail raised and body leaning forward in a playful stance, while the other foxes are gathered close, continuing their dynamic interaction. The overall scene captures the lively and exuberant nature of young foxes at play in their natural habitat.

**<Option>Original Prompt:** Five gray wolf pups frolicking and chasing each other around a remote gravel road, surrounded by grass. The pups run and leap, chasing each other, and nipping at each other, playing.

**Hallucination Detected:**

- Inconsistency with the original prompt.

- Foxes bodies slightly inconsistent with real-world common sense and cause abnormal biological structure. 0.81s

The sudden appearance and disappearance of wolves alternate in the frames, presenting an unreasonable appearance or disappearance. 6.38s ~ 8.97s

**Overall Video Quality Score:** 64.16

**Detection Detail:**

1.key frame and detail frame extraction:

The number of key frame: 3

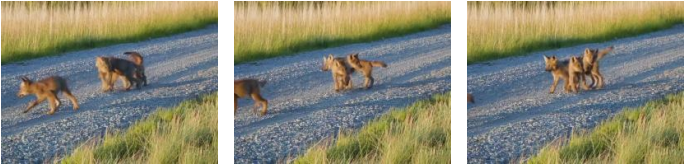

key_frame_1(center): 0.81s
detail_frame_l1(left): 0.16s
detail_frame_r1(right): 1.31s

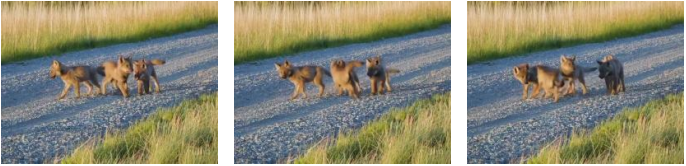

key_frame_2(center): 3.6s
detail_frame_l1(left): 3.11s
detail_frame_r1(right): 3.93s

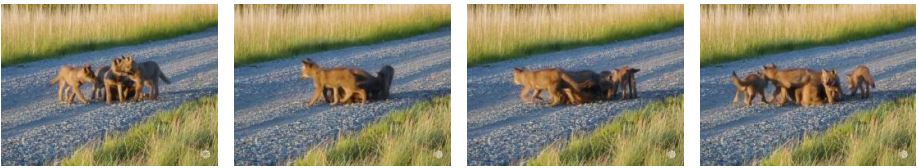

key_frame_3(second): 8.02s
detail_frame_l1(left): 6.38s
detail_frame_r1(third): 8.35s
detail_frame_r2(right): 8.97s

2.object extraction and static KG construction for each key frame:

key_frame_1:{{"objects": ["fox_1", "fox_2", "fox_3"], "relations":[["fox_2", "following", "fox_3"], ["fox_1", "running ahead of", "fox_2"]]}}

key_frame_2:{{"objects": ["fox_1", "fox_2", "fox_3"], "relations":[["fox_1", "next to", "fox_2"], ["fox_2", "next to", "fox_3"]]}}

key_frame_3:{{"objects": ["fox_1", "fox_2"], "relations":[["fox_1", "next to", "fox_2"]]}}

3.object change extraction for each key frame cluster:

key_frame_cluster_1:{"changes":{"fox_1": "unchanged", "fox_2": "unchanged", "fox_3": "unchanged"}}

key_frame_cluster_2{"changes":{"fox_1": "moved forward and left", "fox_2": "moved forward", "fox_3": "initially unchanged, then bounds forward"}}

key_frame_cluster_3:{"changes":{"fox_1": "initial decrease in engagement, stabilizes", "fox_2": "slight separation and reintegrates"}}

4.video summary:

Summary: The video depicts a playful scene featuring three young foxes on a gravel pathway surrounded by grassy areas. The sequence begins with one of the foxes darting across the path, characterized by its motion blur, indicating rapid movement. The next key frame shows the three foxes standing closely together in the middle of the path, playfully interacting with each other, as evidenced by their alert stances and physical proximity. In the final key frame, one fox remains active and engaged, with its tail raised and body leaning

forward in a playful stance, while the other foxes are gathered close, continuing their dynamic interaction. The overall scene captures the lively and exuberant nature of young foxes at play in their natural habitat.

5.inconsistency comparison:
Analysis: Because of the differences in species, number of animals, and nature of the activities described, the inconsistencies substantially affect how viewers might understand and visualize the scene based on the prompt descriptions. These are crucial elements that fundamentally change the depiction of the scene.
Inconsistency Score: 9

6.static hallucination detection:
**key_frame_1:**

| Hallucination Type | Hallucination Analysis | Hallucination Score |
|---|---|---|
| Geometric Structure Irrationality | / | 0 |
| Biological Structure Irrationality | Foxes bodies slightly inconsistent with real-world common sense and cause abnormal biological structure. | 5 |
| Lighting, Shadow and Material Physical Inaccuracy | / | 0 |
| Color Distribution Disharmony | / | 0 |
| Depth of Field and Focal Length Unreality | / | 0 |
| Object Composition and Scene Semantic Inconsistency | / | 0 |
| Motion and Blur Inconsistency | / | 0 |
| Physical Phenomenon Inauthenticity | / | 0 |
| Image Quality Inconsistency | / | 0 |
| Other Hallucination | / | 0 |

**key_frame_2:**

| Hallucination Type | Hallucination Analysis | Hallucination Score |
|---|---|---|
| Geometric Structure Irrationality | / | 0 |
| Biological Structure Irrationality | / | 0 |
| Lighting, Shadow and Material Physical Inaccuracy | / | 0 |
| Color Distribution Disharmony | / | 0 |
| Depth of Field and Focal Length Unreality | / | 0 |

| Object Composition and Scene Semantic Inconsistency | / | 0 |
| Motion and Blur Inconsistency | / | 0 |
| Physical Phenomenon Inauthenticity | / | 0 |
| Image Quality Inconsistency | / | 0 |
| Other Hallucination | / | 0 |

**key_frame_3:**

| Hallucination Type | Hallucination Analysis | Hallucination Score |
| --- | --- | --- |
| Geometric Structure Irrationality | / | 0 |
| Biological Structure Irrationality | / | 0 |
| Lighting, Shadow and Material Physical Inaccuracy | / | 0 |
| Color Distribution Disharmony | / | 0 |
| Depth of Field and Focal Length Unreality | / | 0 |
| Object Composition and Scene Semantic Inconsistency | / | 0 |
| Motion and Blur Inconsistency | / | 0 |
| Physical Phenomenon Inauthenticity | / | 0 |
| Image Quality Inconsistency | / | 0 |
| Other Hallucination | / | 0 |

7.dynamic hallucination detection:

**key_frame_cluster_1:**

| Hallucination Type | Hallucination Analysis | Hallucination Score |
| --- | --- | --- |
| Clipping | / | 0 |
| Implausible Fusion | / | 0 |
| Implausible Appearance or Disappearance | / | 0 |
| Implausible Motion | / | 0 |
| Implausible Transform | / | 0 |
| Implausible Penetration | / | 0 |
| Physical Interaction Errors | / | 0 |
| Logical Interaction Error | / | 0 |
| Other Hallucination | / | 0 |

**key_frame_cluster_2:**

| Hallucination Type | Hallucination Analysis | Hallucination Score |
|---|---|---|
| Clipping | / | 0 |
| Implausible Fusion | / | 0 |
| Implausible Appearance or Disappearance | / | 0 |
| Implausible Motion | / | 0 |
| Implausible Transform | / | 0 |
| Implausible Penetration | / | 0 |
| Physical Interaction Errors | / | 0 |
| Logical Interaction Error | / | 0 |
| Other Hallucination | / | 0 |

**key_frame_cluster_3:**

| Hallucination Type | Hallucination Analysis | Hallucination Score |
|---|---|---|
| Clipping | / | 0 |
| Implausible Fusion | / | 0 |
| Implausible Appearance or Disappearance | The sudden appearance and disappearance of wolves alternate in the frames, presenting an unreasonable appearance or disappearance. | 8 |
| Implausible Motion | / | 0 |
| Implausible Transform | / | 0 |
| Implausible Penetration | / | 0 |
| Physical Interaction Errors | / | 0 |
| Logical Interaction Error | / | 0 |
| Other Hallucination | / | 0 |

8.aggregation and overall scoring:

Overall Hallucination Score Aggregation:

| Inconsistency Score | Static Hallucination Score | Dynamic Hallucination Score |
|---|---|---|
| 9 | 1.11 | 3.35 |

Overall Score: 100 - 2 * 9 - 4 * 1,11 - 4 * 3.35 = 64.16



\end{document}